\documentclass[conference]{IEEEtran}

\usepackage{booktabs} 
\usepackage{verbatim}
\usepackage{subcaption}
\usepackage{pifont}
\usepackage[linesnumbered,ruled,vlined]{algorithm2e}
\SetKwInput{KwInput}{Input}
\usepackage{adjustbox}
\usepackage{amsmath}
\usepackage{tabularx}
\usepackage{multirow}
\SetKwInput{KwInput}{Input}

\usepackage{hyperref}

\newif\ifsubm
\submtrue

\usepackage[colorinlistoftodos, textwidth=2.0cm]{todonotes}

\usepackage{verbatim}

\newcommand{\major}[1]{\ifsubm #1\else \textcolor{blue}{#1}\fi}

\newcommand{\cmark}{\ding{51}}%
\newcommand{\xmark}{\ding{55}}%

\newcommand{\company}{ our cooperation company\xspace}

\def\BibTeX{{\rm B\kern-.05em{\sc i\kern-.025em b}\kern-.08em
    T\kern-.1667em\lower.7ex\hbox{E}\kern-.125emX}}
\begin{document}

\title{A Case Study of Vehicle Route Optimization}

\author{
\IEEEauthorblockN{Veronika Lesch}
\IEEEauthorblockA{\textit{University of Würzburg}\\
Würzburg, Germany \\
veronika.lesch@uni-wuerzburg.de}
\and
\IEEEauthorblockN{Maximilian König}
\IEEEauthorblockA{\textit{PASS Logistics Solutions AG}\\
Aschaffenburg, Germany}
\and
\IEEEauthorblockN{Samuel Kounev}
\IEEEauthorblockA{\textit{University of Würzburg}\\
Würzburg, Germany \\
samuel.kounev@uni-wuerzburg.de}
\and
\IEEEauthorblockN{Anthony Stein}
\IEEEauthorblockA{\textit{University of Hohenheim}\\
Hohenheim, Germany \\
anthony.stein@uni-hohenheim.de}
\and
\IEEEauthorblockN{Christian Krupitzer}
\IEEEauthorblockA{\textit{University of Hohenheim}\\
Hohenheim, Germany \\
christian.krupitzer@uni-hohenheim.de}
}

\maketitle

\begin{abstract}
In the last decades, the classical Vehicle Routing Problem~(VRP), i.e., assigning a set of orders to vehicles and planning their routes \major{has been intensively researched.
As only the assignment of order to vehicles and their routes is already an NP-complete problem,} the application of these algorithms in practice often fails to take into account the constraints and restrictions that apply in real-world applications, the so called rich VRP~(rVRP) and are limited to single aspects.
In this work, we incorporate the main relevant real-world constraints and requirements.
We propose a two-stage strategy and a Timeline algorithm for time windows and pause times, and apply a Genetic Algorithm~(GA) and Ant Colony Optimization~(ACO) individually to the problem to find optimal solutions. 
Our evaluation of eight different problem instances against four state-of-the-art algorithms shows that our approach handles all given constraints in a reasonable time.
\end{abstract}


\section{Introduction}
In the last two decades, the demand for road freight transport increased worldwide; for example, in Germany it increased by 150 billion ton kilometers to around 500 billion ton kilometers~\cite{BibEntry2021Feb}. 
Developments such as increased just-in-time production and online shopping (especially during the Covid pandemic) will further increase those numbers in the next years.
\major{To handle such amount of freight transportation, efficient and correct planning of tours for transports is relevant.
Hence, fast and reliable solutions to the Vehicle Routing Problem~(VRP) are required}.

The classical VRP specifies the assignment of customer orders to vehicles and the optimization of their tours~\cite{golden2008vehicle}, which refers to solving the underlying Traveling Salesman Problem~(TSP).
\major{Tim Pigden stated that the original model of the VRP does not match real-world applications since it does not include concepts of order, separate resources corresponding to the driver, the tractor unit, and the trailer~\cite{Pigden2013}.}
The rich VRP (rVRP) \major{extends this classical VRP by including} additional constraints required for a real-world application, such as pickup and delivery~(PD), time windows~(TW), pause times, trailer capacities, and driver assignments.
Since the rVRP is an NP-complete problem, exact solutions are hard to calculate in \major{short time frames} and, hence, logistic companies often use meta-heuristics to find so-called \textit{good enough} solutions in a reasonable time.
However, \major{due to the complexity,} those approaches do not consider all relevant aspects of the rVRP\major{, i.e., they miss the requirement of multi-objectiveness, and additionally} a manual adjustment \major{to cope with aspects not inherently integrated in the solution} is required.

In this paper, \major{we present the application of nature-inspired algorithms to solve the rVRP within a real-world application software of our cooperation partner, which they use for planning the logistics of their customers.}
Hence, this paper contributes to \major{the rVRP research} by applying nature-inspired algorithms on a multi-objective capacitated VRP with pickup and delivery behavior and time windows.
Our scientific contributions are three-fold:
\begin{itemize}
    \item We define a two-stage strategy for tackling the formulated rVRP including a (i)~VRP-stage that assigns orders to vehicles and a (ii)~TSP-stage that optimizes the tour for each vehicle.
    \item We propose a timeline algorithm within the solving workflow that modifies the planned tours in order to handle time windows and fixed pause stops.
    \item We evaluate our approach in a real-world scenario provided by our cooperation company including eight problem instances with a complexity of up to 100 orders and 13 vehicles.
\end{itemize}

In the following, Section~\ref{sec:relwork} discusses related work.
Section~\ref{sec:problemstatement} defines our problem domain and Section~\ref{sec:approach} presents our approach.
Section~\ref{sec:evaluation} evaluates our approach and discusses the results.
Finally, Section~\ref{sec:conclusion} summarizes the findings of our paper.

\section{Related Work}
\label{sec:relwork}
\major{The TSP and VRP are well-known and highly researched transportation problems that were first mentioned in the last century: the TSP in 1930, and the VRP in 1959.}
Hence, the literature provides many different approaches to both of the problem statements.
Besides the classical VRP, that assigns customer orders to vehicles and optimizes their tours, several extended VRP versions exist.
These versions include additional requirements to the VRP such as capacities of vehicles, time windows, and pickup and delivery behavior. 
Cordeau et al.~\cite{Cordeau2007} introduce a capacity constraint for all vehicles of the fleet that must not be exceeded and thus define the capacitated VRP~(C-VRP)~\cite{WEI2018843,Rabbouch2020}.
In the VRP with time windows~(VRP-TW), each customer order can be defined using additional time windows that refer to opening hours of the location which need to be met by the delivery vehicle~\cite{Cordeau2007,Hintsch2018,Espinoza2016}.
The VRP with pickup and delivery~(VRP-PD) provides the possibility to return goods to depots or transport them from one location to another one and to place multiple pickup and deliveries at one location~\cite{Desaulniers2002,Montero2017Dec,Chavez2016,Shahdaei2016}.
Further, the combination of time windows and pickup and delivery results in the VRP-TW-PD~\cite{Wang2016,Doerner2019,Tchoupo2017,Chami2017}
All versions of the VRP are highly researched and the literature provides a large amount of approaches to tackle these problems.
We are aware, that this summary of related work is only an excerpt and does not provide a complete overview of all relevant literature in this field.
But it represents a spectrum of the main research streams in the area of these specific problems.

\major{In the following, we analyze relevant literature of the last five years that explicitly covers multiple objectives as part of their VRP or TSP approach.
\cite{Tirkolaee2017Robust} addresses multi-trip VRP with intermediate depots and time windows.
The paper provides a robust Mixed-Integer Linear Programming model and addresses the following objectives: travel distances, vehicle costs, and earliness and tardiness penalty costs of services. 
They solve their model using CPLEX, but do not define the time complexity of their problem. 
The authors of \cite{Tirkolaee2019Developing} also address a multi-trip VRP in the domain of urban waste collection. 
They seek to minimize cost objectives, such as traversing costs, employment costs, and exit penalties from permissible time windows. 
Unlike the previous papers, they use Simulated Annealing to solve their problem, but also do not specify the time complexity of their approach.
The advancement of this paper is that they were able to compute near-optimal solutions in less computation time.
In contrast to our work, these two papers do not evaluate their approach using a real-world scenario, but use a randomly generated problem set.
Moreover, they only deal with a limited set of objectives and constraints.}

\major{\cite{dutta2020multi} addresses a multi-objective set orienteering problem using clusters of customers.
The authors assign a predefined profit amount per visit to each customer in a cluster and specify a maximum service time.
Their approach has two objectives: maximizing customer satisfaction and maximizing profit.
The authors use NSGA-II and Strength Pareto Evolutionary Algorithm~(SPEA2), which have a time complexity of $\mathcal{O}(MN^2)$ ($M$ number of objectives, $N$ population size) for NSGA-II and $\mathcal{O}(K^2 log K)$ ($K$ is population size and archive size) for SPEA2.
The advancement of this method is to incorporate customer satisfaction objectives instead of standard cost-based objectives. 
Contrary to this work, we consider each customer as an individual service unit with individual constraints, restrictions, and objectives, and do not consider customer satisfaction metrics.}

\major{A multi-objective model of the capacitated VRP for perishable goods is proposed in~\cite{barma2021multi}.
The objectives of this model are to minimize the quality degradation of goods and to minimize the delivery costs. 
The authors propose an m-ring star distribution network with two types of vehicles and customers, and apply NSGA-II and SPEA2 with the same time complexity statements as in the previously mentioned paper.
The evaluation shows that NSGA-II performs better in terms of quality and costs when using two types for vehicles. 
This work differs from our work in the modelling approach as they use the m-ring star distribution model while we use the two-stage strategy.
Further, they also integrate only a limited number of objectives.}

\major{The authors of \cite{mukherjee2021multi} deal with a multi-objective ring tree problem with secondary sub-depots.
They specify a fixed node as depot and define other primary and secondary sub-depots in combination with three types of customers. 
The objectives include minimizing the total routing cost and minimizing the number of type 3 customers. 
The authors use a discrete multi-objective antlion optimizer with a time complexity of $\mathcal{O}(MN^2)$ ($M$ number of objectives, $N$ population size).
In their evaluation, the authors show that their approach has better efficiency for most test instances.
Contrary to our work, this work focuses on assigning customer orders with basic cost-based objectives, while we apply a variety of real-world objectives.}

\major{Another set of studies focuses on green approaches to VRP variants. 
First, \cite{Tirkolaee2018Hybrid} addresses a multi-trip green capacitated arc routing problem.
The authors aim to minimize the total cost, which consists of routing costs, vehicle costs, and greenhouse gas generation and emission cost. 
They use a hybrid GA with Simulated Annealing for generating initial solutions.
The authors do not specify the time complexity of their approach, but show that their solution performs desirably within a reasonable computation time.
Second, \cite{tirkolaee2020robust} deals with a green VRP with intermediate depots and integrates urban traffic conditions, fuel consumption, time windows, and uncertainty in demands.
They model this problem as robust Mixed-Integer Linear Programming model and solve it using CPLEX.
The integration of urban traffic conditions is a particular advance of this work. 
Third, \cite{ALINAGHIAN2021100802} proposes a Mixed-Integer Linear Programming model for the green inventory routing problem with time windows. 
They attempt to minimize the total cost, which consists of fuel consumption, driver cost, inventory cost, and vehicle cost. 
The authors use an original and an augmented Tabu Search as well as Differential Evolution, but do not specify the time complexity of their approaches. 
In contrast to our work, all the green VRP research approaches focus heavily on integrating green objectives and, hence, include only a limited set of real-world objectives.}

\major{The last set of related works from recent years covers the integration of uncertainty in the pickup demand.
First, \cite{Weber2017Multi} addresses uncertainty in urban waste collection and models the problem as a two-stage multi-objective transportation problem. 
They model uncertainty as grey parameters and apply a procedure to reduce them to real numbers.
They solve their model using revised multi-choice goal programming but do not specify the time complexity of their problem. 
Second, \cite{Weber2017Conic} addresses a multi-choice multi-objective transportation problem and model cost, demand, and supply as multi-choice parameters. 
They reduce their problem to a multi-objective transportation problem by introducing binary variables and applying revised multi-choice goal programming.
However, they do not specify the time complexity of their approach.
Third, \cite{roy2019multi} addresses a multi-objective multi-item fixed-charge solid transportation problem and incorporate fuzzy-rough variables as coefficients of their objective functions and constraints. 
They use a fuzzy-rough expected-value operator to transform the problem into a deterministic one, and apply weighted goal programming and fuzzy programming to find final solutions.
The advancement of this paper is to evaluate and apply it on a real-world case study. 
Fourth, \cite{Tirkolaee2019Robust} also addresses the urban waste collection problem with uncertainties and models the problem as a robust bi-objective multi-trip periodic capacitated arc routing problem under demand uncertainty.
They integrate cost and tour length objectives and solve their problem using CPLEX and a multi-objective invasive weed optimization for real-world problem instances without defining the time complexity.
The particular advance of these approaches is the general applicability of their approaches to model uncertainty.
The main difference between the presented approaches and our approach is that our approach does not deal with uncertainties, but optimizes with fixed predetermined values.
Further, the first three approaches are not evaluated on a real-world data set, but only provide a numerical assessment and sensitivity analyses.
Finally, our approach considers a broader set of objectives compared to the presented approaches.}

In line with the observation of~\cite{Pigden2013}, our analysis of related work shows that existing approaches fail to address the combination of different aspects of the rVRP in such a way that all relevant requirements of a real-world application are taken into account simultaneously.
In this work, we address this research gap and integrate a multitude of real-world requirements.

\section{Foundations on TSP and VRP}
This section presents most important foundations on the TSP and VRP.
\subsection{Traveling Salesman Problem}
\label{sec:found:tsp}
The TSP is a highly researched optimization problem, first mentioned in the 1830s~\cite{voigt1831handlungsreisende}. 
It deals with the problem of finding the shortest possible route from a given initial city that visits all other cities exactly once and then returns to its initial city.
This problem is known to be NP-hard and, hence, there is probably no algorithm that solves this problem in polynomial time. 

The first philosophical mentions of the TSP are found in a literary magazine as a book advertisement in the 1830s~\cite{voigt1831handlungsreisende}. 
The advertised book describes the daily life of a traveling salesman and gives instructions on how to do the job, hints on good routes through Germany and Switzerland, and suggests places to stay.
The first mathematical focus is found in the 1950s with the definition proposed by Merrill M. Flood~\cite{flood1956traveling}.
Later, it was implicitly proven that the TSP is NP-hard when Richard M. Karp proved the NP-hardness of the Hamiltonian cycle~\cite{karp1972reducibility}.
The TSP has been and is still being extensively researched, as it can be used for a variety of real-world applications.

The TSP aims to find minimal routes within a network of cities and can therefore be represented by graphs.
This graph consists of nodes representing cities and edges representing paths between nodes. 
The distances between nodes are represented by the weights of edges.
The TSP is a minimization problem where the goal is to find a path that visits all existing nodes.
This path must start from a particular node, which serves as the start and end node. 
In addition, the TSP can also be modeled mathematically by Integer Linear Program formulations first proposed by Dantzig, Fulkerson, and Johnson~\cite{dantzig1954solution} and Miller, Tucker, and Zemlin~\cite{miller1960integer}. 

Exact approaches to solving the TSP include brute-force algorithms, branch-and-bound techniques~\cite{balas1983branch}, and linear programming~\cite{dantzig1954solution,miller1960integer}.
Since the TSP is provably NP-hard, its complexity does not allow the computation of exact solutions for large problem spaces. 
Therefore, the focus has shifted to heuristic approaches that compute feasible solutions in a short time.
David S. Johnson and Lyle A. McGeoch provide a comprehensive overview of heuristic approaches, including greedy algorithms, nearest neighbor, k-opt, Tabu Search, Simulated Annealing, ACO, and Neural Networks, in their book chapter~\cite{johnson20188}. 
Besides, ACO are also commonly applied to the TSP~\cite{DORIGO199773}.

\subsection{Vehicle Routing Problem}
\label{sec:found:vrp}
The VRP is a generalization of the previously introduced TSP, which was first introduced in 1959~\cite{dantzig1959truck}.
While the TSP considers a single vehicle for which a route must be planned, the VRP considers multiple vehicles for which routes must be planned taking into account a number of customers and orders. 
The goal is to reduce the total driving distance for all vehicles while serving all customers. 
Since it is a generalization of the NP-hard TSP, the VRP is also an NP-hard problem~\cite{karp1972reducibility}.
In 1964, Geoff Clarke and John W. Wright proposed the first effective greedy heuristic for computing feasible solutions for the VRP~\cite{Clarke1964}.

Similar to the TSP, the VRP can also be represented as a graph consisting of nodes and edges modeling customers and paths between customers, respectively.
Again, the edge weight represents the distance between two nodes, and the goal is to minimize travel distances.
However, unlike planning a single route with all nodes in the TSP, the VRP must compute a set of routes that consider all customers.
The number of routes to be planned is equal to the number of vehicles.
Three different mathematical models for the VRP are proposed in the literature according to~\cite{letchford2006projection}: (i)~Vehicle flow formulation~\cite{laporte1992vehicle}, (ii)~Commodity flow formulation~\cite{gavish1978travelling}, and (iii)~set partitioning formulation~\cite{letchford2006projection}. 

The survey by Gilbert Laporte and Yves Nobert categorizes exact solution approaches to the VRP into three groups~\cite{laporte1987}: (i)~Direct tree search, (ii)~dynamic programming, and (iii)~Integer Linear Program.
However, the VRP is an NP-hard problem and, so there is no algorithm that finds an exact solution in polynomial time.
This is the reason why researchers and practitioners use heuristic approaches such as Savings Algorithm, Tabu Search, Simulated Annealing, ACO, ACO, and hybrid approaches~\cite{toth2014vehicle}.

The basic definition of the VRP is often extended by further constraints, which are necessary for the application to real use cases~\cite{toth2014vehicle}.
The capacitated VRP considers capacities of vehicles that must not be exceeded.
This leads to the assumption of homogeneous or heterogeneous fleets, where in the homogeneous case all vehicles are assumed to have the same capacity, which is not the case in the heterogeneous instances. 
While the standard variant of VRP considers only one depot, the pickup and delivery variant~(P\&D) of VRP allows considering multiple pickup and delivery locations.
This means that the delivery of a good must be scheduled with the same vehicle that picks up the good.
This constraint leads to the consideration of multiple depots within one VRP, while the base version considers exactly one depot that is both the origin and destination.
In addition, uncertainty can be part of the definition of VRP, since parts or all of the input may be unknown at the time of planning.
This leads to dynamic VRP models that are able to handle uncertainty at design time~\cite{pillac2013review}.

\section{Problem Statement and Complexity}
\label{sec:problemstatement}
\begin{figure*}[htb]
    \centering
	\includegraphics[width=0.7\textwidth]{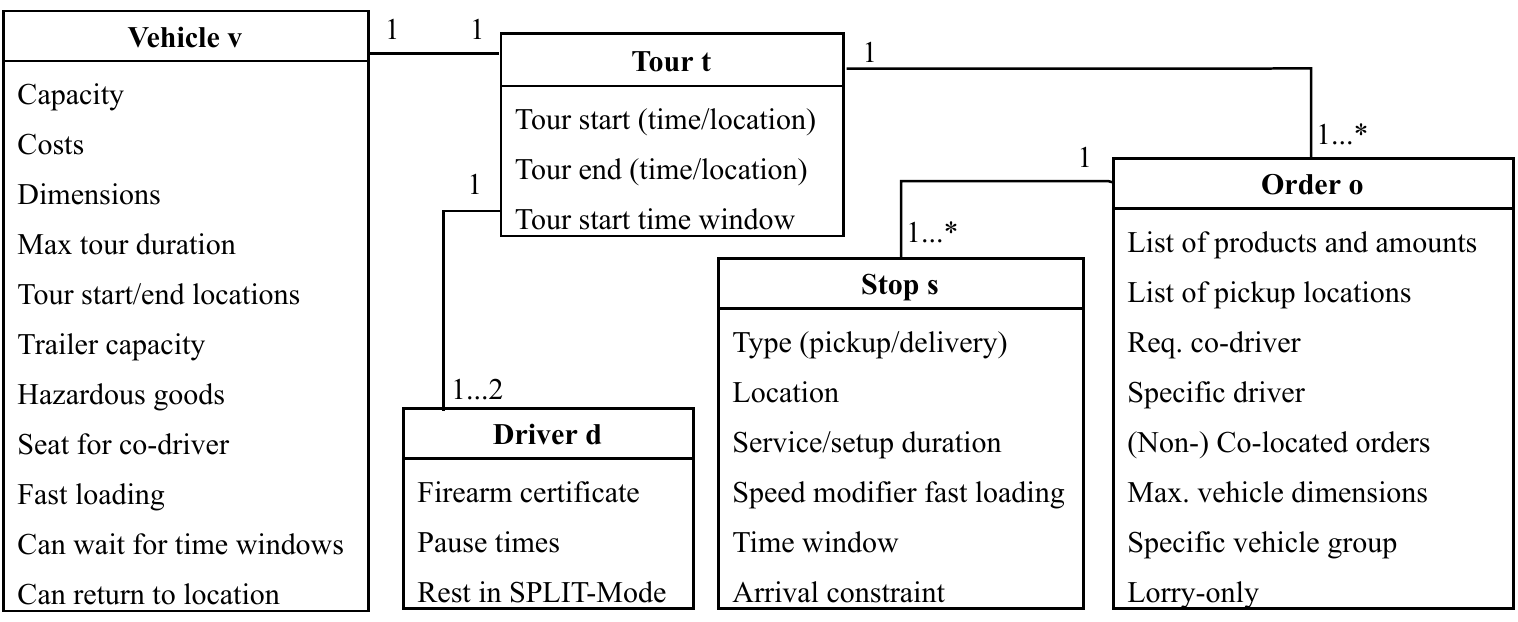}
    \caption{Domain model of the rich VRP addressed in this paper.}
    \label{fig:domainmodel}
\end{figure*}

We cooperate with a consultancy company which plans the logistics operations for many different customers and therefore has a diverse set of requirements which we define in the following.
Figure~\ref{fig:domainmodel} illustrates the considered version of a rVRP as domain model. 
We define a \emph{tour} $t_j$ as the assignment of customer orders $o \in O$ to vehicles $v \in V$ and drivers $d \in D$: $t_j = (v_i,D_j,O_j)$ with a set of drivers $D_j \subseteq D$ and a set of orders $O_j \subseteq O$ assigned to tour $t_j$ driven with vehicle $v_i$.
The central goal is to find a set of tours~$T = \{t_j\}$ so that all orders are assigned while minimizing the cost function defined in Section~\ref{sec:cost_function}.
A tour has a tour start and a tour end, each specified by a time and a location. 
The tour start time window defines the range in which the tour must start. 

A \emph{vehicle} $v_i$ is always assigned to one tour $t_j$. 
Each vehicle has a capacity, described by the number of pieces, volume~($m^3$), or weight~(kg).
The costs for using a vehicle are defined per hour, per kilometer, per tour, or per stop on the tour.
The height, width, length~(cm), and weight~(kg) of a vehicle define its dimensions. 
Each vehicle has a maximum tour duration, after which it must be at the tour end location. 
The tour start/end locations represent a list of locations at which the tour can start and a list of locations at which the tour can end.
The option for a trailer specifies whether the vehicle can pull a trailer.
Since special properties of a vehicle are required for carrying hazardous goods, we integrate the possibility to specify whether a vehicle has the ability to carry those goods or not. 
This enables a valid assignment of orders to vehicles even for hazardous goods.
The fast-loading property states whether the vehicle can load and unload faster compared to other vehicles. 
This property influences the required service and setup time and, hence, the time required at a specific stop.
It provides the possibility to switch to a vehicle with fast-loading property to match time windows.
A vehicle might provide the possibility to wait for a specific time window at a given stop and might be allowed to return to a stop multiple times.

Furthermore, up to two \emph{drivers} $D_j$ are determined for each tour $t_j$.
For some orders, a co-driver is required for loading and unloading bulky goods.
Drivers might require a special training or certificate, for example, a firearm certificate is required for cash transports.
By specifying the required certificates for drivers with regards to a specific order, a valid assignment of drivers to tours is possible within the optimization, making an additional post-processing step unnecessary.
Legislation might prescribe a fixed set of pause times for each driver.
Sometimes it may be possible to schedule pause times within a service, such as within a pickup or a delivery (called SPLIT-Mode). 
Otherwise, pause time need to be scheduled while driving in between two stops.

Additionally, customer \emph{orders} $O_j$ need to be serviced during a tour $t_j$. 
Each order contains a list of products with the amount specified by quantity in pieces, weight~(kg), or volume~($m^3$).
For each order, one can specify one or more pickup locations, a co-driver requirement, or the assignment of a specific driver.
Orders that should or should not be delivered in the same tour can be defined as a list of (non-) co-located orders.
Maximum vehicle dimensions, a vehicle from a specific vehicle group, or a lorry-only service can be required.

For each order, at least one \emph{stop} $s$ needs to be scheduled.
Each stop is either a pickup or a delivery stop and has a specific location.
Hence, we usually schedule two stops for an order first, to pick the goods up from a storage facility and a delivery stop at the customer location.

The setup/service duration contains the time the vehicle stands still while loading or unloading.
The speed modifier for fast loading vehicles defines the saved time during the setup/service duration if this stop is assigned to a vehicle with this property.
Each stop has a time window that specifies at which interval the driver needs to arrive or finish the service.

\major{We now present a mathematical definition for the capacitated VRP using the vehicle flow formulation~\cite{dantzig1954solution} as originally applied on the VRP by~\cite{laporte1992vehicle}.
Let $G=(V,A)$ be a graph where $V={1,\dots, n}$ is a set of vertices representing cities, or in our case customers, with the depot located at vertex 1.
$A$ is a set of arcs~$(i,j)$ with $i\neq j$ that are associated with a non-negative distance matrix $C = (c_{ij})$.
This distance matrix can be interpreted as the travel distance between the vertices or as travel costs of this arc.
At the depot, a set of $m$ homogeneous vehicles with capacity $D$ are available with $m_L \leq m \leq m_U$.
Let $x_{ij} (i\neq j)$ be a binary decision variable that is equal to 1 iff the optimal route contains arc $(i,j)$. }
\major{
\begin{align}
    \text{minimize}\qquad &\sum_{i\neq j} c_{ij} x_{ij} \label{eq:min}\\
    \text{subject to}\qquad &\sum_{j=1}^n x_{ij} = 1 \qquad (i = 1,\dots, n),\label{eq:c1}\\
    &\sum_{i=1}^n x_{ij} = 1 \qquad j = 1,\dots, n),\label{eq:c2}\\
    &\sum_{i,j \in S} x_{ij} \leq \mid S \mid - v(S) \label{eq:c3} \\ 
        &\qquad (S \subset V\backslash\{1\}; \mid S\mid \geq 2),\nonumber\\
    &x_{ij} \in \{0,1\}\label{eq:c4}\\
        &\qquad (i,j = 1,...n; i\neq j)\label{eq:c4}\nonumber.
\end{align}
}
\major{In this formula, Equation~\ref{eq:min} forms the minimization equation to minimize the distances of all routes.
Constraints~\ref{eq:c1} and \ref{eq:c2} ensure that all vertices are visited exactly once and that exactly one vehicle arrives and departs from this vertex. 
Constraint~\ref{eq:c3} forms the sub-tour elimination constraint with $v(S)$ being an appropriate lower bound on the number of required vehicles for this problem.
Finally, Constraint~\ref{eq:c4} forms the integrity constraint defining $x_{ij}$.}

\major{This formula models a standard VRP mathematically. 
However, the problem addressed in this paper handles a wide variety of constraints and restrictions that are not modeled above. 
Since a full mathematical modeling would go beyond the scope of this paper and would unnecessarily lengthen it, we briefly outline the concepts that can be applied to model our problem.
The requirements of a homogeneous fleet, individual vehicle capacities, capabilities of vehicles regarding fast loading and transporting of hazardous goods, as well as fixed costs per vehicle can be addressed by introducing an additional set of vehicle types $B=1,\dots, b$ similar to~\cite{kecceci2021mathematical}.
To include the decision for one or two drivers, and an optional trailer, three indices can be added to the decision variable $x_{ijkl}^{ab}$ where $l$ represents the optional trailer and $a$ and $b$ represent the drivers. 
Driver requirements can be added analogously to the heterogeneous vehicles by introducing sets of driver types.
The pickup and delivery requirement including the possibility of multiple pickup locations is also introduced in~\cite{kecceci2021mathematical} which requires additional decision variables for pickup and delivery demand~($p_i, d_i$) as well as variables that summarize the loaded pickup and delivery load at each vertex~($z_{ij}, t_{ij}$). 
To fulfill all order requirements such as (non) co-located orders, special vehicle restrictions, whether the stop can be planned using the SPLIT mode as well as time windows per stop, these factors can be easily integrated as individual constraints.
Finally, the tour start and end time, the start time window, the maximum tour duration, and a set of pause times can be integrated by adding variables that count the required time per tour as well as start and end times per stop. 
As already mentioned, we do not want to go into more detail about the mathematical model, but want to refer the interested reader to~\cite{zirour2008vehicle,martello2011surveys}.
}

\major{In the final paragraphs of this section, we analyze the problem space and define the complexity of the addressed problem statement. 
Let $V$ be the number of vehicles and $O$ the number of orders.
Every order contains at least one pickup and one delivery stop, and hence, the number of stops to be assigned is \mbox{$2 \cdot O$}.
Additionally, the options for start and end location of each vehicle as well as required pause times are included as stops.
However, as tour start and end location options as well as the number of pause times are fixed constants these do not increase the problem complexity in the $\mathcal{O}$ notation.
Since multiple options are possible for each stop, a virtual vehicle called \textit{slack} contains the unused options for all stops.
Theoretically, every stop can be assigned to every vehicle including the slack which results in $(2 \cdot O)^{(V+1)}$ possible distributions of stops to vehicles. 
In addition to the distribution problem, the sequence of stops is relevant for optimizing the TSP and every assignment of $S$ stops to one vehicle has $S!$ possible sequences.
Hence, for every distribution of stops to vehicles, $V$ different TSPs need to be solved. 
The sequence on the slack is irrelevant since it is a virtual vehicle and, therefore, does not need to be considered in the optimization.
To summarize the complexity of the problem, we use a chain representation where a VRP solution is represented by a chain containing all stops and $V+1$ indices, where the chain is cut to distribute its parts to the vehicles.
This representation is feasible since the individual TSP chains are independent of each other.
Hence, at least $(2 \cdot O)!$ TSP solutions and $2 \cdot O$ cut indices exist and the complexity of the problem can be defined as follows:
\begin{equation}
Complexity = (2 \cdot O)! \cdot 2 \cdot O \in \mathcal{O}(O! \cdot O)
\end{equation}}

\major{In summary, this section first introduces a detailed software-engineering based definition of our addressed VRP. 
Further, we provide the mathematical model of a VRP using the vehicle flow formulation and summarize possibilities to model all constraints addressed in this work.
Then, we discuss the problem complexity and show that the problems have superpolynomial complexity with regards to their input size.
Both VRP and TSP are proven to be NP-complete problems and the required time to solve these problems for all known algorithms is superpolynomial in the input size.
Approaches to solving NP-complete problems tend to be limited to approximation, randomization, restriction, parametrization, and heuristics. 
Due to the often large input size of VRPs, often meta-heuristics are applied as they may provide a sufficiently good solution within reasonable time.
Since our work was done in cooperation with a consulting company, we defined the requirement to deliver a solution within a few seconds to minutes in addition to the already mentioned requirements for the solution. 
The high complexity of the problem in combination with the requirement of a fast time-to-solution led us to selecting meta-heuristics as solution approaches: (i) GA as a prominent evolutionary approach often applied in VRP use cases, and (ii) ACO as representative of nature-inspired, particle swarm optimization.
With this selection, we aim to test as diverse solutions as possible and at the same time be able to meet all our requirements with at least one of them.
}

\section{Approach}
\label{sec:approach}
This section describes our approach for tackling the \major{multi-objective optimization of the} rVRP and presents the overall complexity of our approach. 
The complexity discussion of each part of the overall complexity can be found in the according sections.
We introduce our two-stage strategy and \major{present the Timeline approach for the time windows and pause stops as well as the cost function with which we address the multi-objective problem.}
\major{The two-stage strategy reduces the overall complexity for the optimization mechanisms by dividing the solution space into two individual problems. 
We cannot make firm statements about the impact on optimality, however, we assume that our approach does not negatively impact optimality; especially in light of the fact that the meta-heuristics applied are already non-optimal.
The Timeline approach further reduces complexity by offloading compliance with the specified time windows to a third separate step.
This complexity reduction allows the algorithms to yield valid results already after a few seconds optimization time and further optimize the results significantly within the first minutes.
For a detailed analysis of the advantages introduced by the approaches of this section, we refer the interested reader to our evaluation in Section~\ref{sec:evaluation}.
Finally, we introduce our six-score priority cost function which enables the applied approaches to handle the multi-objective properties of the problem.}

\begin{figure}[htb]
\centering
	\includegraphics[width=\columnwidth]{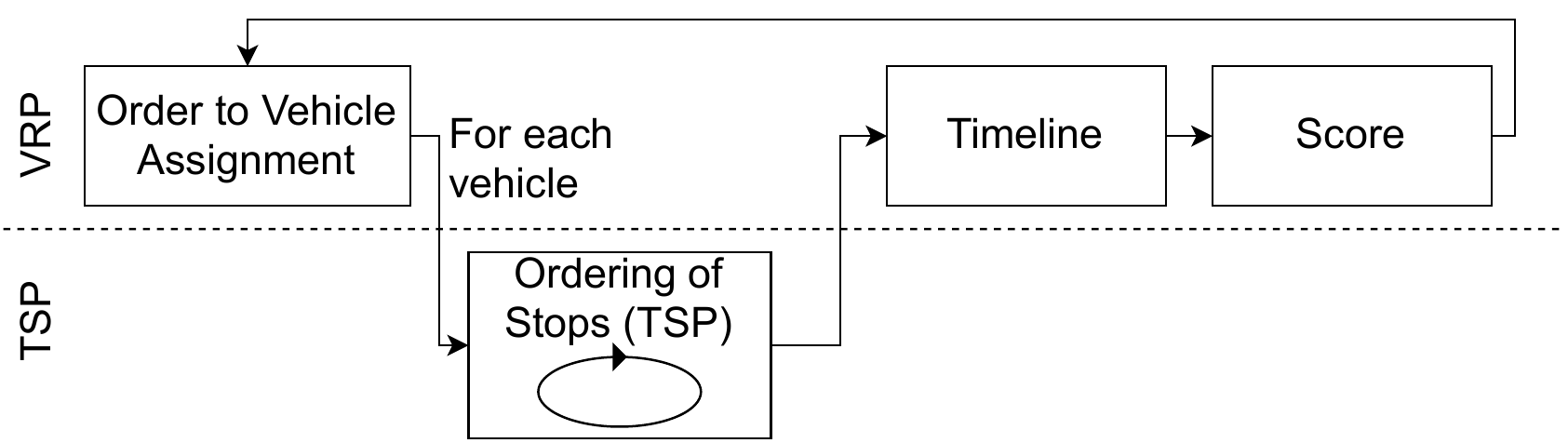}
    \caption{Overview of the two-staged strategy.}
    \label{fig:2stages}
\end{figure}

Due to the high complexity of the problem, and inspired by~\cite{Chen2010}, we divide the problem into two stages as depicted in Figure~\ref{fig:2stages}.
First, we address the problem of distributing all orders, including pickup and delivery options, to the available vehicles~(VRP-stage).
In this step, several assignment-related constraints such as order restrictions are addressed.
However, many of the above mentioned constraints are sequence-dependent and, hence, the nested TSP instance for each vehicle needs to be solved. 
In the TSP-stage, the TSP-solver starts and solves an individual TSP instance for each vehicle. 
We retrieve the actual stop-to-stop route from a route planning service of\company{} and plan the order of stops at this stage.
The solved TSP instances are then sent back to the VRP-stage that performs our Timeline algorithm presented in Section~\ref{sec:timeline}.
to determine time windows and pause stops for each tour. 
Finally, the distribution can be rated with regards to the cost function explained in the next section and the algorithm decides whether the current distribution should be kept or discarded.
Depending on the size of the VRP and the nested TSP instances, either exact (works for smaller problem spaces) or heuristic approaches (for large problem spaces) can be used to solve the stages. 
Since we do not want to restrict applicability of our proposed system to only work for small TSP instances, we propose our heuristic approaches based on GA and ACO for both stages.
\major{This two-stage strategy summarizes our overall approach to tackle our rVRP.
The complexity of our approach can be derived by summarizing the above mentioned operation steps depicted in Figure~\ref{fig:2stages} into a runtime notation:}
\major{
\begin{align*}
    T_{overall}(n) &=  T_{vrp}(n) + m \cdot T_{tsp}(n_2) \\
    &+ m \cdot T_{timeline}(n_2) + T_{cost}(n)
    \label{eq:overall}
\end{align*}
}
\major{with $n$ being the number of stops to plan, $n_2$ being the number of stops per tour, $m$ being the number of required trucks in the solution. 
Hence, this formula summarizes the complexity of the VRP algorithm for all given stops, the complexity of the TSP and the complexity of the timeline algorithm multiplied by the number of trucks, and the complexity of the score calculation.
The overall $\mathcal{O}$ notation can be derived by inserting the mentioned $\mathcal{O}$ complexities of all parts from the according sections.}

\subsection{Timeline Algorithm for Time Windows and Pause Stops}
\label{sec:timeline}
In this section, we introduce the \emph{timeline algorithm} to match the pause times and fit as many time windows of stops as possible (see Algorithm~\ref{algo:timeline}).
This algorithm fits all pause times first and then tries to fulfill all time window requirements.
All stops are shifted to fulfill all pause times and hence, no pause time violations will be present after the execution of this algorithm.
\begin{algorithm}
	\KwInput{Sequence of stops $seq$}
	Initialize timeline $t$ and place pause times\\
	Buffer = tour start interval\\
	Penalty = 0\\
	\ForEach{$s$ in $seq$}{ 
		Add $s$ as early as possible on $t$\\
		\If{TW not yet reached}{
			Shift to right until TW is met or buffer is empty\\
			Reduce buffer by shifted seconds
		}
	\If{(buffer is empty AND TW not yet reached)}{
		Add waiting time to $t$ until TW starts}
		\ElseIf{TW already passed}{
			Increase penalty by missed seconds   
		}
	}
	\Return Penalty
	\caption{Pseudo-code for the Timeline Algorithm.}
	\label{algo:timeline}
\end{algorithm}

This algorithm iterates over each sequence of stops (i.e., once per vehicle and tour) and calculates a penalty for the score.
It first initializes the timeline with given start and end times of a tour, pause times, and time windows.
Then, it iterates over the sequence of stops and places all stops as early as possible taking into account the sequence, tour start interval, and its time windows.
If the current timestamp is too early for the pause time or time window's starting time, the algorithm shifts the whole chain of stops (excluding the pauses) to a later starting point while keeping all previous time windows and the tour start interval.
This also includes a recalculation of all previously placed stops by a defined amount of time regarding the start time of each stop.
In case a shift is not possible, the vehicle waits until the time window for this stop starts. 
In case a pause time is reached while driving from one stop to another, the algorithm adds a pause on the route.
If the SPLIT mode is activated, that is, a pause during services is possible, the pause time is added to the service time. 
If it is deactivated, the full service needs to be shifted to after the pause.
After the placement of all stops with time windows and the placement of pause times on the timeline, the scores $H_3$ to $S_3$ are recalculated and fed back to the VRP-stage of the algorithm to judge the quality of the solution.
\major{Since this algorithm iterates once over all stops per sequence, that is, all stops per vehicle, the complexity of this algorithm can be summarized as:}
\major{
\begin{equation}
    T_{timeline}(n_2) \in \mathcal{O}(n_2)
\end{equation}
}

\subsection{Cost Function}
\label{sec:cost_function}
Since the rVRP addressed in this paper exhibits a high diversity of constraints and restrictions, we propose to use a cost function consisting of six priority scores for the evaluation of the generated solutions.
\major{These six scores address all objectives of the problem defined in Section~\ref{sec:problemstatement} and enable the applied approaches to handle this multi-objective problem.}
The six scores are divided into three hard scores $[H_1, H_2, H_3]$ and three soft scores $[S_1, S_2, S_3]$.
The hard scores assess the solution's feasibility and, hence, are handled as hard constraints, while the soft scores represent the solution's quality.
The cost function is designed to form a minimization goal for the optimization process.

In case the capacity of vehicles and trailers is exceeded, the first hard score $H_1$ sums up the exceeded capacity by subtracting the defined vehicle capacity~($v_{cap}$) from the planned vehicle capacity~($v_{pcap}$).
Further, it adds a value of 100 score points for each fault in existing order restrictions such as a co-driver requirement~($f_{or}$) and a violation in order dependencies like co-located orders~($f_{od}$) where the operator \# indicates the number of violations. 
We decided to use the multiplier 100 to balance the impact of an order restriction violation (counted as number of violations) compared to a capacity exceed (counted as difference of weight, volume, or the like).
\major{The calculation for this score has a complexity of $\mathcal{O}(n)$ as it requires to iterate over all stops.}

\begin{equation}
    H_1 = \sum_{v \in V} max((v_{pcap} - v_{cap}),0) + 100 \cdot \# f_{or} + 100 \cdot \# f_{od}
\end{equation}
The second hard score $H_2$ deals with the pickup and delivery order, the entry order, and the vehicle assignment.
First, it is checked whether the pickup is done prior to the corresponding delivery with 100 score points for each fault~($f_{pd}$).
Afterward, the vehicle-specific tour start and end locations are examined and one score point is added for each fault~($f_{se}$).
Finally, this score evaluates whether all stops that require a specific vehicle are serviced by such a vehicle~($f_{sv}$) and whether all planned returns to stops are allowed~($f_{sr}$).
Any fault adds one score point to $H_2$. 
\major{Similar to $H_1$, this score also requires to iterate over all stops and check the requirements and, hence, has a complexity of $\mathcal{O}(n)$.
However, this score can also be calculated during the iteration over all stops for the first hard score, and, thus, does not increase the overall complexity of the score calculation.}
\begin{equation}
    H_2 = \sum_{v \in V} 100 \cdot \# f_{pd} + \# f_{se} + \# f_{sv} + \# f_{sr}
\end{equation}

The hard score $H_3$ sums the seconds the tour duration~($t_{dur}$) exceeds the maximum duration~($t_{maxdur}$) and the planned tour end~($t_{pend}$) exceeds the end constraint~($t_{end}$). 
\major{This score iterates over all vehicles and calculates time restrictions and, hence, has a complexity of $\mathcal{O}(m)$.
Since the number of vehicles ($m$) is constant and significantly smaller than $n$ this calculation can be assumed to be constant in terms of the overall score complexity in $\mathcal{O}$ notation.}
\begin{equation}
    H_3 = \sum_{v \in V} max((t_{dur} - t_{maxdur}),0) + max((t_{pend} - t_{end}),0)
\end{equation}

The soft scores assess the quality of the solutions.
The first soft score $S_1$ assesses how good the solution matches each time window~($tw$) in the set of predefined time windows~($TW$).
It sums up how many seconds the planned time window~($tw_p$) exceeds the given time window~($tw_g$). 
Therefore, the seconds the planned time window starts~($tw_{p,s}$) ahead of the given time window are calculated and added to the seconds the planned time window ends~($tw_{p,e}$) after the given time window ends.
\major{For the calculation of this score, all stops need to be assessed and, hence, the complexity of this calculation can be summarized as $\mathcal{O}(n)$.
Again, this calculation can be integrated into the calculation of the fist two hard scores and does not increase the overall complexity.}
\begin{align*}
    S_1 &= \sum_{v \in V} \sum_{tw \in TW} max((tw_{g,s} - tw_{p,s}),0) \\
    &+ max((tw_{p,e} - tw_{g,e}),0)
\end{align*}

The second soft score $S_2$ summarizes driven kilometers~($dist$), waiting~($ time_{wait}$), driving~($ time_{drive}$), and service times~($ time_{service}$).
The individual values can be multiplied by the costs per vehicle, trailer, and personnel to represent the costs for a tour.
Since these objectives form the main goal of the defined VRP in this work, we decided to integrate them into one score and, hence, assign the same priority to these objectives.
\major{Similar to the previous score, all stops need to be assessed and the complexity can be summarized as $\mathcal{O}(n)$.
Again, this does not increase the overall complexity.}
\begin{equation}
    S_2 = \sum_{v \in V} dist + time_{wait} + time_{drive} + time_{service}
\end{equation}
The last soft score $S_3$ refers to the delay of a driver starting his/her tour~($t_{pstart}$) after the defined start~($t_{start}$), the number of visited locations~($loc$), and the chain length~($cl$), that represents the number of stops to be serviced during the tour.
This score integrates further soft constraints that are less important than the main objective goals in $S_2$ and is only assessed if several solutions perform equally well on $S_2$.
Hence, this score is used to decide which solution performs best, if multiple solutions perform equally well in our main objective score $S_2$.
\major{To calculate this score, constants for each vehicle need to be summarized and the complexity is $\mathcal{O}(m)$ which can be seen as constant and does not increase the overall complexity.}
\begin{equation}
    S_3 = \sum_{v \in V} max((t_{start} - t_{pstart}),0) + \# loc + cl
\end{equation}

To save computation time, we only calculate the scores $H_3$, $S_1$, $S_2$, and $S_3$ if the previous hard scores are down to zero.
Otherwise the solution is considered to be infeasible, i.e., the hard scores are not down to zero.
Since we implemented our score system as priority scores that need to be minimized, the first smaller value of a score level---starting at $H_1$ and ending at $S_3$---decides which of the two solutions performed better.

An example score value for a VRP solution that meets all capacity constraints, breaks one order restriction and sticks to all entry order and location-specific constraints can look like $\text{Hard }[100, 0, 0]$, $\text{ Soft }[120, 2919200, 1235]$. 
The $H_1$-value of 100 represents the order restriction fault of this solution, while $H_2$ and $H_3$ have a value of 0 indicating, that these constraints are all met.
The $S_1$-value of $120$ means that the vehicles of this solution break time windows by 120 seconds.
The $S_2$-value $2919200$ is the sum of all service, driving, and waiting times, while the last score~($S_3$) refers to the delay of starting times and the number of visited tours.

\major{In summary, we define the complexity of the cost function as:}
\major{
\begin{equation}
    T_{cost}(n) \in \mathcal{O}( 4 \cdot n + 2 \cdot m) = \mathcal{O}(n)
\end{equation}
}
\major{We argue that the number of vehicles~($m$) is a constant since the number of available vehicles is fixed and significantly smaller than the number of orders which results in $\mathcal{O}(n)$.}

\section{Genetic Algorithm}
\begin{figure}[htb]
\centering
	\includegraphics[width=\columnwidth]{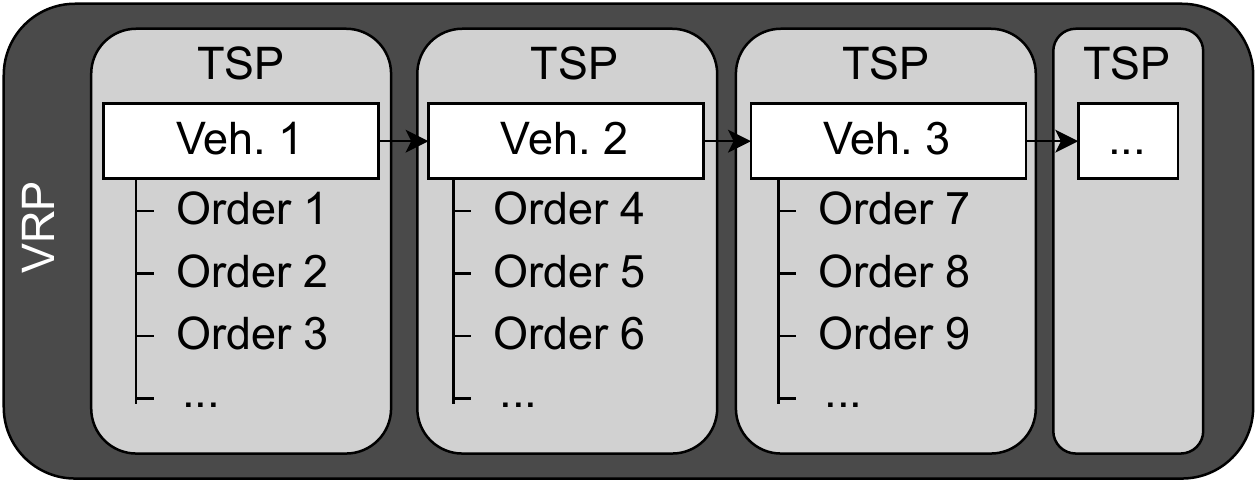}
    \caption{Illustration of an object-oriented genome representation of the GA.}
    \label{fig:gagenome}
    \vspace{-0.2cm}
\end{figure}

We now describe the genetic algorithm~(GA) we applied, inspired by the approach in~\cite{holland1992genetic}. 
Figure~\ref{fig:gagenome} presents the genome representation used for the GA.
Each genome contains a set of vehicles each representing a TSP instance.
Each vehicle holds a list of orders. 
This list is passed to the TSP-stage that determines the most beneficial ordering of this list.
\major{We define the complexity of our GA individually for both stages in their according subsections.}

\subsection{VRP-stage}
\major{We define the complexity of the VRP-stage of the genetic algorithm based on the population size~$P$, the number of generations~$G$, the crossover probability~$Prob_c$, the complexity of the crossover~$T_{c,vrp}^{ga}$, the mutation probability~$Prob_m$, and the mutation complexity~$T_{m,vrp}^{ga}$.
The complexity of the TSP-stage, the Timeline approach and the cost function are already included in the complexity of the overall approach and thus, they do not need to be added here.
Again, $n$ is the number of orders, and $m$ represents the number of vehicles.}
\major{
\begin{equation}
    T_{vrp}^{ga}(n) = P \cdot G \cdot Prob_c \cdot T_{c,vrp}^{ga}(n) + Prob_m \cdot T_{m,vrp}^{ga}(n) 
\end{equation}
}
\major{Since the parameters $P$, $G$, $Prob_c$, $Prob_m$, and $m$ are constants, the complexity of the VRP-stage of the GA can be reduced to:}
\major{\begin{equation}
    T_{vrp}^{ga}(n) = T_{c,vrp}^{ga}(n) + T_{m,vrp}^{ga}(n) 
\end{equation}}

\major{In the following, we present the process of our adapted VRP-stage GA.
Each iteration of the VRP-stage GA performs four steps after a population initialization phase:} (i)~breed new individuals, (ii)~solve the TSP-stage for each vehicle, (iii)~calculate the score, and (iv)~maintain population size.
These steps are repeated until either a predefined number of unimproved iterations or a given computation time is reached as summarized in Algorithm~\ref{algo:gavrp}.
\begin{algorithm}
	\KwInput{Orders $o$, vehicles $v$, max. iterations $i_{max}$, population size, mutation probability, crossover probability}
	Create initial individual and mutate it to initialize population\\
	\While{unimproved iterations ($ui$) $<$ $i_{max}$ AND current runtime $<$ max. runtime}{
	    \While{population size $<$ 2 $\cdot$ initial population size}{
		    Select two individuals\\
		    Apply a random crossover operator\\
		    Apply a random mutation operator with probability $p$\\
		    \ForEach{$v$}{
		      Solve TSP-stage for offspring
		    }
		    Apply Timeline algorithm to match time windows\\
		    Add offspring to the population
		    }
		Sort population by score and remove worst half\\
		\If{improved best score}{$ui$ = 0}
		\Else{$ui++$}
	}
	\Return population
	\caption{VRP-stage pseudo-code for the GA.}
	\label{algo:gavrp}
\end{algorithm}

Instead of initializing the population purely at random, the GA creates an initial individual by assigning orders to vehicles with regards to vehicle restrictions or co-location requirements.
Afterward, the remaining orders are matched to the vehicles based on stop-to-stop distances, that is, a stop that has the minimum distance to already assigned stops of a vehicle is assigned to this vehicle.
This solution is improved by iterating over all stops and moving them to other vehicles in order to improve the average stop-to-stop distances for all vehicles.
Then, for each required individual as specified in the population size, the GA selects one mutation operator from the list of available operators randomly and applies it to the individual to create the whole initial population.
For each individual in this population and for each vehicle, the TSP-stage solves the stop sequence.
Afterward, the Timeline algorithm is applied, and the already introduced score is calculated for each individual and used as its fitness value.

After the initialization phase---the creation of the initial population~(line~1)--- the VRP-stage GA iterates until one of the above mentioned termination criteria is met~(line~2).
Each iteration, that is, each generation, breeds new individuals until the population size has doubled~(line~3).
Therefore, the algorithm randomly selects uniformly distributed from three possible selection operators (that are introduced later) to breed a new individual from two parent individuals~(line~4):
(i)~select two individuals randomly based on a uniform distribution;
(ii)~select two individuals randomly based on a predefined probability, where the individual with the best score has the highest probability;
and (iii)~a tournament selection where ten solutions compete pair-wise and the winner is selected for recombination.
Then, the algorithm randomly selects a crossover operator from the set of provided operators and applies it on this pair of individuals~(line~5).
Afterward, the algorithm mutates the new individual with a probability $p_{vrp}=0.5$ using a randomly selected mutation operator~(line~6).
With defining a set of selection and mutation operations and their random selection in each population, we cope with the variety of constraints and aim at a higher diversity in the population.

For each newly created individual, the algorithm forwards the TSP instances to the TSP-stage that solves this instance and returns ordered lists of stops~(lines 7 and 8).
Then, the algorithm applies the Timeline algorithm to match the given time windows~(line~9).
Finally, the algorithm calculates the score of the new individual and adds it to the current population~(line~10).
Since, the population size doubled during this iteration, half of the population needs to be discarded to match the predefined population size~(line~11).
Therefore, the algorithm sorts the population according to the achieved score and removes the worst half of individuals.
This affects the next generation of the algorithm as only the best performing individuals are kept for recombination in the next iteration and therefore accelerates the convergence of the GA.

The crossover operators use two individuals for breeding a new offspring. 
Therefore, chains or parts of chains are copied from the parent individuals to the new individual.
The remaining stops, that is the sub chains that are not copied to the new individual, are assigned based on the stop-to-stop distance of each vehicle regarding already assigned stops.
Since our problem definition includes diverse constraints, we define the following three crossover operators to breed new individuals in multiple ways to increase the diversity of the population:
\begin{enumerate}
	\item  The \textit{OverlapCrossover} operator copies stops located on both parents to the new individual. Remaining stops are added to the vehicle of the new individual with lowest distance to existing stops of this vehicle. \major{$\mathcal{O}(n)$}
	\item The \textit{ScoreBasedCrossover} operator copies the chain of the parent with lower costs to the new individual. The remaining stops from the other parent are added similar to the first crossover. \major{$\mathcal{O}(n)$}
	\item The \textit{SelectionCrossover} operator selects one of the parents randomly and assigns the chain to the new individual. The remaining stops from the other parent are added similarly to the other crossovers. \major{$\mathcal{O}(n)$}
\end{enumerate}
\major{Considering that these operators are applied on pairs of individuals, we define the overall complexity of the crossover computation as:}
\major{
\begin{equation}
    T_{c,vrp}^{ga}(n) \quad\in \mathcal{O}(\frac{n}{2} * n) \quad\in \mathcal{O}(n^2)
\end{equation}}

Mutation operators are used for breeding new individuals from a single parent individual and increasing the diversity of the population.
For each individual that should be mutated, we select one mutation operator randomly.
Since it is not guaranteed that a mutation operator produces a valid individual we restart the mutation with another randomly selected operator in case the individual is invalid.
Since our problem definition includes diverse constraints, we define the following mutation operators, each modifying the genome in a different way, aiming at a specific constraint.
By providing this diverse set of mutation operators, we deal with the variety of constraints and are able to keep the diversity of the population as high as possible rather than focusing on a single mutation operator.
\begin{enumerate}
	\item The \textit{ClearVehicleMutator} removes all stops of a random vehicle and assigns them to other vehicles, based on a location and distance-based rating. \major{$\mathcal{O}(n)$}
	\item The \textit{SwapVehicleMutator} swaps chains of two different vehicles, excluding the vehicle's start and end locations. Since this operator is applied at the VRP-stage consisting of multiple vehicles and their assigned orders, it is considered a mutation. \major{$\mathcal{O}(1)$}
	\item The \textit{OutlierMutator} iterates through every vehicle's stop chain, selecting the stop pair that contributes most to the distance-based rating and moving it to another vehicle. \major{$\mathcal{O}(n)$}
	\item The \textit{MoveOrderMutator} takes up to three orders of one vehicle and moves them to another vehicle, based on the distance rating. This behavior is repeated for a random number of times with a maximum of four times. \major{$\mathcal{O}(n)$}
	\item The \textit{CloseToOtherVehicleChainMutator} selects a stop from a chain close to another chain, and moves the order for this stop to the nearby chain. \major{$\mathcal{O}(n)$}
	\item The \textit{SavingsMutator} iterates over all stops of every vehicle and computes the highest saving of distance when moving one order to another vehicle. Additionally, predecessors and successors are moved to another vehicle if this reduces the distance. This mutator avoids overlapping tours. \major{$\mathcal{O}(n)$}
\end{enumerate}
\major{We define the overall complexity of the mutation computation as:}
\major{
\begin{equation}
    T_{m,vrp}^{ga}(n) \quad\in \mathcal{O}(n^2)
\end{equation}}

\subsection{TSP-stage}
\label{sec:ga_tsp}
\major{Analogously to the algorithm complexity of the VRP-stage, we define the complexity of the TSP-stage as:}
\major{\begin{equation}
    T_{tsp}^{ga}(n_2) = T_{c,tsp}^{ga}(n_2) + T_{m,tsp}^{ga}(n_2) + T_{cost}(n_2)
\end{equation}}
\major{Since we included the computation of the cost function for the VRP-stage explicitly in the overall complexity $T_{overall}$ we do not need this complexity in the VRP-stage.
On the contrary, this computation complexity is not included for the TSP-stage and we need to include it explicitly in $T_{tsp}^{ga}(n_2)$.}

The TSP-stage of the algorithm calculates the sequence and options (i.e., the list of possible locations) selection for each vehicle independently.
Hence, the following description always captures performed steps for the tour of a single vehicle.
At the beginning, the initial population is created similarly to the initialization of the VRP-stage by calculating a first valid individual.
For this individual, the algorithm starts with a random stop and assigns the remaining stops based on the stop-to-stop distances, that is, the algorithm selects always the nearest stop compared to the last assigned stop.
Then, the algorithm mutates this individual by applying randomly selected mutation operators to create the required amount of individuals for the initial population. 

After the initialization phase, the TSP-stage GA performs similar steps compared to the VRP-GA.
It iterates until the maximum amount of unimproved iterations are executed and breeds new individuals until the population size has doubled in each iteration.
For the new individuals, the algorithm selects and recombines two randomly chosen parent individuals using a random crossover operator.
Afterward, the algorithm mutates the individual with a certain mutation probability $p_{tsp}=0.5$ and a randomly selected operator and adds it to the population.
As the population size is doubled, the algorithm omits the worst half of the population to accelerate the convergence.

Again, the crossover operators combine two parents into one new individual.
We define the following three crossover operators to breed new individuals in different ways and keep the diversity of the population high.
The crossover operators in the TSP-stage are inspired by~\cite{Hussain2017}: 
\begin{enumerate}
	\item The \textit{RandomCrossover} randomly chooses the next possible stop from the beginning of the parents' chain while removing stops already contained in the offspring. \major{$\mathcal{O}(n_2)$}
	\item The \textit{OrderedCrossover} performs a classical two-point crossover and combines the genome of both parents. \major{$\mathcal{O}(n_2)$}
	\item The \textit{PartiallyMappedCrossover} works similar to the OrderedCrossover but assigns the remaining stops outside the interval at the beginning of the chain based on the indices of their parents which is the main difference to the one in the literature. \major{$\mathcal{O}(n_2)$}
\end{enumerate} 
\major{In line with the complexity in the VRP-stage, we define the overall complexity of the crossover computation at the TSP-stage as:}
\major{
\begin{equation}
    T_{c,tsp}^{ga}(n_2) \quad\in \mathcal{O}(\frac{n_2}{2} * n_2) \quad\in \mathcal{O}(n_2^2)
\end{equation}}

Additionally, we define the following operators concerning the TSP-stage, inspired by related work~\cite{Hussain2017}.
Again, we decided to provide a diverse set of mutators and select random ones in each iteration to increase the diversity of the population.
\begin{enumerate}
	\item The \textit{ReverseMutator} reverses the sequence of all successive pickup and delivery pairs. \major{$\mathcal{O}(n_2)$}
	\item The \textit{SimpleMoveMutator} moves one stop to another feasible position in the chain, taking into account the constraint of pickup-delivery order. The TourBegin and TourEnd nodes are protected and omitted. \major{$\mathcal{O}(1)$}
	\item The \textit{SimpleSwapMutator} swaps the positions of two stops on the chain. \major{$\mathcal{O}(1)$}
	\item  The \textit{MultiOptMutator} combines the previous two mutators and applies the SimpleMoveMutator or the SimpleSwapMutator up to three times. \major{$\mathcal{O}(1)$}
	\item  The \textit{NeighborhoodSwapMutator} is similar to the SimpleSwapMutator, but it works based on distance improvement when swapping stops. It tries all possible swaps in the chain for one random stop and performs the swap with the highest distance improvement. \major{$\mathcal{O}(n_2)$}
	\item The \textit{SavingsTSPMutator} selects the stop that produces the highest saved distance when moving it in the chain. The delta of the distance concerning the whole chain is calculated and the highest distance savings move is executed. \major{$\mathcal{O}(n_2^2)$}
	\item The \emph{OptionsMutator} selects a random stop with at least one option and randomly replaces it with one of the other possible options. \major{$\mathcal{O}(1)$}
	\item The \emph{OptionsChainMutator} rotates the options for the whole chain and replaces all stops with a possible option of this stop. \major{$\mathcal{O}(n_2)$}
\end{enumerate} 
\major{In line with the complexity in the VRP-stage, we define the overall complexity of the mutation computation at the TSP-stage as:}
\major{
\begin{equation}
    T_{m,tsp}^{ga}(n_2) \quad\in \mathcal{O}(n_2^2)
\end{equation}}

This section introduced our domain-adapted GA \major{and presented complexity definitions. }
First, we presented our object oriented genome presentation used and proposed two stages of this algorithm.
Then, for each stage of the algorithm, we provide a domain-specific set of crossover and mutation operators that are randomly chosen in each offspring computation.
These operators enable the algorithm to cope with the various constraints included in this work and aim at maintaining a high diversity of the population. 

\section{Ant Colony Optimization}
This section explains the developed two-staged ACO algorithm inspired by~\cite{dorigo1996ant}.
We modified the classical ACO algorithm for both stages to accommodate for the complexity of the rVRP:

\begin{itemize}
    \item We replaced the pheromone initialization by a heuristic one concerning the actual stop-to-stop distances to kick-off the optimization from the first step onward.
    \item We use a deterministic ACO in the VRP-stage, this means that we start with an assignment of stops to vehicles based on the pheromone matrices. This helps to decrease bad performing solutions at the start.
    \item The stops for pickups and deliveries are assigned in pairs, so that one vehicle needs to serve both stops in one tour. This prevents creating invalid solutions that put pickup and deliveries on different vehicles. 
\end{itemize}

\subsection{VRP-Stage}
\label{ch:approach:sec:aco_distribution}
\begin{figure}[htb]
\centering
	\includegraphics[width=0.8\columnwidth]{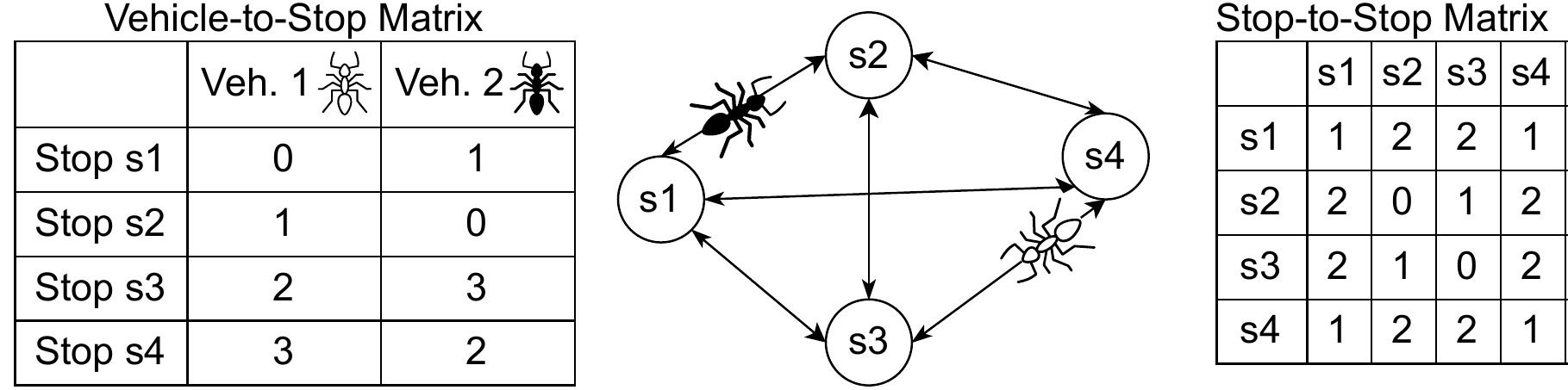}
    \caption{Graph representation of the VRP problem for the ACO algorithm.}
    \label{fig:acomodel}
    \vspace{-0.3cm}
\end{figure}

Similar to the VRP-stage of the GA, the VRP-stage of the ACO algorithm assigns stops to vehicles and optimizes the solutions. 
The assignment of stops to vehicles and its optimization works with two pheromone matrices as illustrated in Figure~\ref{fig:acomodel}, where each ant represents one vehicle.
The \emph{vehicle-to-stop} matrix represents the occupied capacity of vehicles so that ants select the vehicles with enough free space first.
The algorithm updates this matrix after each assignment with the current available space of the according vehicle.
We performed preliminary tests using a single pheromone matrix which showed us that this value is not enough to determine a good order to vehicle distribution.
Instead, the stops that are already assigned to a vehicle have further influence on the final solution as a good clustering of stops per vehicle seems to be advantageous.
Hence, we introduce the \emph{stop-to-stop} matrix that covers the distance between stops and is used to determine the next stop to be added.
By implementing the second matrix, stops with a close distance to each other are more likely to be assigned to the same vehicle:
First, an ant selects a stop based on stop-to-stop matrix that is reachable from its current location and has the shortest distance.
Then, the ant searches for vehicles that have enough space for this order.
We then assign a probability of selecting each of these vehicles by adding the vehicle-to-stop pheromone value~(available space) and the stop-to-stop pheromones to all already assigned stops of this vehicle~(stop-to-stop distances).
Based on these probabilities, the ant selects a vehicle randomly. 
This means, the higher the amount of aggregated pheromones, the better the vehicle suits this order, the higher the probability to select this vehicle.

Algorithm~\ref{algo:acovrp} summarizes the behavior of the ACO algorithm using the two pheromone matrices in the VRP stage.
\begin{algorithm}
    \KwInput{Stops $s$, Vehicles $v$, max. iterations $mi$}
    Initialize pheromone matrices\\
	\While{unimproved iterations ($ui$) $<$ $mi$ AND current runtime $<$ max. runtime}{
		Assign $s$ to $v$ based on matrices\\
		\ForEach{$veh$ in $v$}{
		    Solve TSP-stage
		}
		Apply Timeline algorithm to match time windows\\
		Calculate score of this solution\\
		Try add it to set of best solutions\\
		Update pheromones\\
		Evaporate pheromones \\
		\If{current solution better than best solution}{$ui$ = 0}
		\Else{$ui++$} 
	}
	\Return best solutions
	\caption{VRP-stage pseudo-code of the ACO.}
	\label{algo:acovrp}
\end{algorithm}
First, the pheromone matrices are initialized with the apriori knowledge of vehicle capacities and stop-to-stop distances~(line~1).
Additionally, an empty set of solutions is initialized in which the best solutions are stored.
The number of stored solutions is defined as twice the number of vehicles of a specific problem instance.
We decided to double the vehicle number to have at least one ant per vehicle and a second ant for a further optimization round.
Then, a loop starts iterating until a maximum number of iterations that were not able to improve the solution quality are executed~(line~2). 
In each iteration, one ant is placed at the graph and assigns all stops to the vehicles with regards to both pheromone matrices~(line~3).
In order to keep the idea of Novelty Search~\cite{lehman2008exploiting} and avoid getting stuck in local optima, a small amount of distributions are created probabilistic.
Afterward, the algorithm passes a TSP instance per vehicle to the TSP-stage of the ACO which optimizes its sequence~(lines~4 and 5). 
The returned TSP instances are then passed to the Timeline algorithm to match time windows~(line~6).
Afterward, the algorithm calculates the final scores for this solution~(line~7).
Then, the algorithm updates the pheromone matrices using the scores of the solutions in the set and performs a pheromone evaporation step with a probability of 5\% which we identified in a preliminary parameter study~(lines~9 and 10).
If the found solution is better than the worst one in the solution set, or the solution set is not yet full, the solution is added to this set~(lines~11 to 14).
If the solution is better than the best solution so far, the number of unimproved iterations is reset to zero.
Otherwise, the number of unimproved iterations is incremented.
Afterward, the next iteration starts, another ant is placed at the graph, and assigns the stops to vehicles.

\major{
Derived from the algorithm we define the complexity of our VRP-stage ACO using the complexity of the initialization~$T_{init,vrp}^{aco}(n)$, the maximum of the number of unimproved iterations~$ui$ and the maximum runtime, the complexity of the TSP-stage~$T_{tsp}^{aco}(n_2)$ in combination with the number of vehicles~$m$, the complexity of the pheromone update~$T_{upd,vrp}^{aco}(n)$ and the pheromone evaporation~$T_{evap,vrp}^{aco}(n)$.
The complexity of the TSP-stage, the Timeline approach and the cost function are already included in the complexity of the overall approach and thus, they do not need to be added in this complexity definition.
}
\major{
\begin{align}
\begin{split}
    T_{vrp}^{aco}~ &= ~T_{init,vrp}^{aco}(n) \\ 
    &+ max(ui, \text{max. runtime}) \cdot ( m \cdot T_{tsp}^{aco}(n_2)) \\
    &+ T_{upd,vrp}^{aco}(n) \\ 
    &+ T_{evap,vrp}^{aco}(n)
\end{split}
\end{align}
}
\major{The initialization computation complexity considers the initialization of the vehicle-to-stop and the stop-to-stop matrices. 
While the vehicle-to-stop matrix is initialized with zero values, its complexity is $\mathcal{O}(1)$.
The stop-to-stop matrix contains the stop-to-stop distances and requires iteration over all stops which results in $\mathcal{O}(n^2)$.
In summary, we define the initialization complexity as:
}
\major{
\begin{equation}
    T_{init,vrp}^{aco}(n) \in \mathcal{O}(n^2)
\end{equation}
}

The matrix update in this stage works with a comparison of the score value to the last best and worst scores.
Due to the fact, that the algorithm deals with a multi-level priority score, that is any broken constraint in level $i$ is more important than any improvement in level $i+1$, we decided to include this knowledge in the pheromone update strategy.
This way, we want to provide more weight for higher score levels than to lower score levels and direct the search of the algorithm to improve the convergence speed.
As summarized in Equation~\ref{eq:pheromones}, the new pheromones for every score level~$i$ are calculated by multiplying the score factor~$f_i$ with a pheromone base value~$p_i$, divided by the score level~(one for~$H_1$, two for~$H_2$, and so on) to give more weight to the more important scores.
We distinguish two cases to set $p_i$: if the current score is better than the worst score ever found, we set  $p_i=1$; if the current score is worse than the worst score, we set $p_i=0.25$.
By this, we give the pheromones of reasonable solutions more weight than of bad ones and hope to gain a faster improvement of the found solutions since many more non-feasible solutions exist.
The already mentioned idea of integrating Novelty Search brings the possibility of worse solutions than the currently worst one.
\begin{equation}
\label{eq:pheromones}
\text{pheromones} = \sum_{i}\frac{f_i \cdot p_i}{i}
\end{equation}

Equation~\ref{eq:fscore} shows the calculation of the score factor~$f_i$.
The variable~$ws_i$ refers to the current worst score, $bs_i$~to the current best score, and $s_i$~to the current score value of the respective level~$i$.
By using this formula, we decrease the pheromone amount of solutions with lower scores than the current worst score and exponentially award better solutions.
\begin{equation}
\label{eq:fscore}
f_i = \left\lvert \frac{(ws_i - s_i) ^ 3 }{(ws_i - bs_i)^3}\right\rvert
\end{equation}

\major{The complexity of the pheromone update includes the number of score levels, the number of stops and the actual calculation of the new score. 
While the number of score levels is a constant value of six, the number of pheromones to be updated is defined as $n^2$.
The actual calculation of the update pheromone value can be done in $\mathcal{O}(1)$.
This results in an overall pheromone update complexity of:
}
\major{
\begin{equation}
    T_{upd,vrp}^{aco}(n) \in \mathcal{O}(n^2)
\end{equation}
}

\major{The pheromone evaporation complexity depends on the number of pheromone values to be updated, which is $n^2$ and the complexity of the actual evaporation computation.
Since we use a fixed evaporation factor in this work, the complexity of the evaporation computation can be reduced to:}
\major{
\begin{equation}
    T_{evap,vrp}^{aco}(n) \in \mathcal{O}(n^2)
\end{equation}
}

\subsection{TSP-stage}
\label{ch:approach:sec:aco_order}
The TSP-stage works with a single \emph{stop-to-stop} pheromone matrix representing the probabilities, that is the distance to move from one stop to another.
The diagonal values refer to the probability of a stop to be the first stop taking the vehicle's start locations into account. 
The other values represent the probabilities to move from one stop to another.
We initialized this matrix again with knowledge about the stop-to-stop distances and hence, represent the actual distance between the stops from the first iteration onward instead of an equal initialization which would require some time to converge to the actual distances.
However, it might happen that order dependencies, order restrictions, or time windows require another stop sequence than shortest first, so we decided to maintain a small probability for every stop.
The algorithm starts iterating and places one ant at any location in the graph in every iteration.
The ant then decides---depending on the column for the current stop containing the values to every other stop---which stop to visit next.
We add a visibility feature to the matrix to guide the ant in a way to first select the pickup stop and Afterward the delivery stop.
Hence, we set the visibility of a delivery stop to false if the ant did not pickup the products for this order beforehand and the ant cannot see this stop.
This aims at further reducing the convergence time of the algorithm.
After one ant finished its walk and returned with a sequence of stops, the algorithm calculates the score for this sequence.
Afterward, the algorithm updates the pheromones similar to the update procedure in the VRP-stage and evaporates the pheromones with a probability of 5\%. 
Further, we apply the principle of Elitism---i.e., the matrix is additionally updated with the current and global best solutions so far---to improve the solution quality even more~ (cf.~\cite{Catay2009}). 
This behavior guides the algorithm to search for better solutions in the neighborhood of already good solutions.
The TSP-stage iterates until a maximum number of unimproved iterations occurred, the maximum runtime is exceeded or the path of the ants converged, that is, all ants select the same path. 

\major{Analogously to the algorithm complexity of the VRP-stage, we define the complexity of the TSP-stage. 
However, we need to add the complexity of the cost function computation as this is not part of the overall complexity defined earlier.}
\major{
\begin{align}
\begin{split}
    T_{tsp}^{aco}(n_2) &= T_{init,tsp}^{aco}(n_2)\\
        &+ T_{upd,tsp}^{aco}(n_2)\\
        &+ T_{evap,tsp}^{aco}(n_2) \\
        &+ T_{cost}(n_2)
\end{split}
\end{align}
}

\major{The initialization computation complexity solely considers the initialization of the stop-to-stop matrix as this stage works with a single matrix. 
Again, the stop-to-stop matrix contains the stop-to-stop distances and requires iteration over all stops which results in a complexity of:
}
\major{
\begin{equation}
    T_{init,tsp}^{aco}(n_2) \in \mathcal{O}(n_2^2)
\end{equation}
}

\major{Since the pheromone update in the TSP-stage works analogously to the one in the VRP-stage, the complexity can be similarly defined as:
}
\major{
\begin{equation}
    T_{upd,tsp}^{aco}(n_2) \in \mathcal{O}(n_2^2)
\end{equation}
}

\major{Finally, the pheromone evaporation factor is a constant value that needs to be assigned to all values in the stop-to-stop matrix of this stage.
This results in a complexity of:}
\major{
\begin{equation}
    T_{evap,tsp}^{aco}(n_2) \in \mathcal{O}(n_2^2)
\end{equation}
}

\section{Evaluation}
\label{sec:evaluation}
First, we present our evaluation methodology in Section~\ref{sec:methodology}, where we define the problem instances, algorithms we use for comparing our proposed approaches, evaluation procedure, and \major{algorithm parameterizations.}
Then, we present our evaluation results in Section~\ref{sec:evalResults}\major{, derive implications for practitioners in Section~\ref{sec:implications},} and discuss threats to validity in Section~\ref{sec:threats}.

\subsection{Evaluation Methodology}
\label{sec:methodology}
\major{In this section, we first introduce the real-world database and define the problem instances we derived to use them for our evaluation.
We then present alternative algorithms that we use as reference values in the evaluation.
Further, we present the parameterization of our algorithms and summarize them in Table~\ref{tab:params}.
Finally, we present the methodology we use to evaluate our approaches.}

Since we handle a real-world rVRP, we decided to use a real database for our evaluation instead of a benchmark instance since we require a huge level of detail for each order, vehicle, and driver.
This would force us to adjust the available benchmark instances which would reduce the comparability of the results what is the main advantage of these instances.
Therefore,\company{} provided a database of real VRPs containing 30 vehicles with different costs, capacities, and capabilities, 15 matching trailers with different specifications, and 30 drivers that can be assigned to vehicles with different capabilities. 
Further, the database contains three depots and 450 orders with according locations around the German city Stuttgart. 
Unfortunately, we are not allowed to make this dataset publicly available since it is part of a non-disclosure agreement.

From this set of data, we define eight different problem instances for evaluating our proposed algorithms.
In line with our separated handling of TSP and VRP instances, we decided to first evaluate the TSP-stage isolated and Afterward apply the algorithms on the VRP-stage that includes solving nested TSP instances.
For the evaluation of the TSP-stage, we define three problem instances:
(i)~a small problem instance of ten orders without pickup and delivery~(PD) and pause times~(TSP-I),
(ii)~a large problem instance of 30 orders without PD and pause times~(TSP-II),
and (iii)~the large problem instance of 30 orders without PD but with pause times~(TSP-II-P).
We similarly define three problem instances for evaluating the VRP-stage:
(i)~a small problem instance of 53 orders and 5 vehicles without PD and pause times~(VRP-I),
(ii)~the small problem instances combined with pause times~(VRP-I-P),
and (iii)~a large problem instance of 100 orders, 13 vehicles without PD and pause times~(VRP-II).
Since we did not include PD behavior, that is, each order has differing pickup and delivery stops, in the previous problem instances, we add two further instances that require PD behavior:
(i)~a TSP problem instance with ten orders, one vehicle with PD but without pause times~(TSP-PD)
and (ii)~a VRP problem instance with 62 orders, seven vehicles with PD and pause times~(VRP-PD).
In all problem instances, time windows are given for orders and need to be handled by the algorithms. 
However, pause times are only integrated if we explicitly stated it, that is, in the problem instances TSP-II-P, VRP-I-P, and VRP-PD.  
Using the real-world data explains the unusual amount of orders and vehicles since the minimum required vehicles depend on the characteristics of the orders.
The extension \emph{P} of the problem instance label indicates that for this problem instance we add the following pause times: 9:30-10:00~AM, 11:30~AM-12:00~PM, and 2:30-3:00~PM.
We here only consider static pause times to evaluate the ability of our algorithms to fulfil this requirement.
However, also flexible pause times can easily be included to replace the static ones.

\begin{table}[htb]
\centering
\caption{Overview of the evaluated Problem Instances~(PI).}
\label{tab:probleminstances}
\begin{adjustbox}{width=0.7\columnwidth}
\begin{tabular}{|l|r|r|c|c|}
\toprule
\textbf{PI} & \textbf{Orders} & \textbf{Vehicles} & \textbf{P/D} & \textbf{Pause Times} \\ \midrule
TSP-I & 10 & 1 & \xmark & \xmark \\ \midrule
TSP-II & 30 & 1 & \xmark & \xmark \\ \midrule
TSP-II-P & 30 & 1 & \xmark & \cmark \\ \midrule
VRP-I & 53 & 5 & \xmark & \xmark \\ \midrule
VRP-I-P & 53 & 5 & \xmark & \cmark \\ \midrule
VRP-II & 100 & 13 & \xmark & \xmark \\ \midrule
TSP-PD & 10 & 1 & \cmark & \xmark \\ \midrule
VRP-PD-P & 62 & 7 & \cmark & \cmark \\ \bottomrule
\end{tabular}
\end{adjustbox}
\end{table}

We compare the performance of our algorithms (GA, ACO) against four alternative algorithms.
Since\company{} provides several algorithms for comparison that are already implemented in OptaPlanner, we decided to also use OptaPlanner for an easy comparison of our new implementations. 
Hence, we implement our algorithms in the OptaPlanner Framework (cf. \url{https://www.optaplanner.org/}) using version \textit{7.31.0.Final}.
Nevertheless, it is of course possible to implement our approach without OptaPlanner.

Table~\ref{tab:algocomp_t} provides essential information on the functional requirements supported by each compared algorithm.
First, we apply a deterministic Brute Force algorithm provided by OptaPlanner that supports all requirements of our scenario.
Since this complete and optimal algorithm requires high computation time, it is only applied to the smallest test instance.
The second algorithm is based on a Savings algorithm~\cite{Clarke1964} and used by\company{} in cases with a homogeneous vehicle fleet, a single depot, and no pickup and delivery problem.
Even if we know on which approach this algorithm is based, we call it Blackbox-I as we have no insight into the details of the implementation.
The third algorithm (Blackbox-II) is an extension to the above mentioned Blackbox-I algorithm covering a multi-depot problem and more complex pause time rules.
Both Blackbox algorithms are proprietary algorithms developed by\company{}.
The fourth algorithm supports all features required for our rVRP as it uses our model of the problem inside OptaPlanner and is an implementation of Tabu Search~\cite{glover1986future} provided by default from the OptaPlanner's Local Search~(LS) algorithms.

Since we modelled the rVRP inside OptaPlanner additional optimization could be applied such as exhaustive search, hyperheuristics or partitioned search. 
However, we decided to use the Tabu Search implementation as promising representative of Local Search algorithms.
Further, other optimization techniques could be applied on the rVRP such as exact algorithms by using an adjusted penalty function.
However, several restrictions of our problem statement prevented us from using these as for example we retrieve the stop-to-stop distance from a service of\company{} and this information is not available as fixed adjacency matrix that could be transferred to other algorithms easily. 
Further, the integration of all handled constraints into one penalty function is problematic since diverse constraints need to be reduced to one value.  
This single value is not a good indicator which of the constraints is violated and therefore a directed search towards an optimum is hardly possible.

\begin{table*}[htb]
\centering
\caption{Overview on the compared algorithms and their capabilities with respect to the requirements of the rVRP.}
\label{tab:algocomp_t}
\begin{tabular}{|l|c|c|c|c|c|c|}
\toprule
Capabilities & Brute Force & Blackbox-I & Blackbox-II & Local Search & GA & ACO \\ \midrule
Capacities & \cmark & \cmark & \cmark & \cmark & \cmark & \cmark \\\midrule
Setup Times & \cmark & \xmark & \xmark & \cmark & \cmark & \cmark \\\midrule
Time Windows & \cmark & \cmark & \cmark & \cmark & \cmark & \cmark \\\midrule
Tour Start Time Window & \cmark & \cmark & \cmark & \cmark & \cmark & \cmark \\\midrule
Order Restrictions & \cmark & \cmark & \cmark & \cmark & \cmark & \cmark \\\midrule
Fixed Pause Times & \cmark & (\cmark) & \cmark & \cmark & \cmark & \cmark \\\midrule
Heterogeneous Fleet & \cmark & \xmark & \xmark & \cmark & \cmark & \cmark \\\midrule
Multiple Depots & \cmark & \xmark & \cmark & \cmark & \cmark & \cmark \\\midrule
Pickup/Delivery & \cmark & \xmark & (\cmark) & \cmark & \cmark & \cmark \\\midrule
Stop Options & \cmark & \xmark & \xmark & \cmark & \cmark & \cmark \\\midrule
Allow Return & \cmark & \xmark & \xmark & \cmark & \cmark & \cmark \\ \bottomrule
\end{tabular}
\end{table*}

We evaluate our two proposed algorithms against the alternative four algorithms on all previously defined problem instances.
The probabilistic algorithms (LS, GA, ACO) are executed 30 times with different random seeds to deliver representative results for comparison.
We summarize the parametrization of our algorithms in Table~\ref{tab:params}.
For defining the population size of the genetic algorithm, we use statistics of the problem instance to be solved.
Therefore, we use 10\% of the number of orders, add the number of vehicles to the power of 1.25, the number of pickup delivery orders, and twice the number of stops containing multiple options. 
We derived these values in a preliminary parameter study and focused on providing a reasonable number of individuals regarding the complexity of the problem.
We set the mutation probability to be 50\%, hence, on average half of the newly created individuals are mutated and set the termination criterion to a maximum iterations without improvement to be 500. 
For the ACO we use an evaporation factor of 0.05, the size of the set of best solutions so far to ten, and the maximum number of iterations without improvement to 500.

\begin{table}[htb]
\centering
\caption{\major{Definition of algorithm-dependent parameters used for the evaluation.}}
\label{tab:params}
\begin{adjustbox}{width=0.99\columnwidth}
\begin{tabular}{|l|c|}
\toprule
\multicolumn{2}{|c|}{\textbf{Genetic Algorithm}}\\ \midrule
Population size & $0.1 \cdot \#o + \#v^{1.25} + \#\text{(PD-orders)} + 2 \cdot \#\text{(multi-option stops)} $ \\ \midrule
Mutation probability ($p_{vrp}$, $p_{tsp}$) & 0.5\\ \midrule
Max. unimproved iterations & 500 \\ \midrule \midrule
\multicolumn{2}{|c|}{\textbf{Ant Colony Optimization}}\\ \midrule
Evaporation factor & 0.05\\ \midrule
Size set of best solutions ($N$) & 10 \\ \midrule
Max. unimproved iterations & 500\\ \bottomrule
\end{tabular}
\end{adjustbox}
\end{table}
For the evaluation runs, we used a server exclusively for our measurements with the following specifications:
Two Intel(R)Xeon(R) CPU E5-2667 v4 processors with 3,20~GHz each with 16~GB of RAM. 
Windows Server 2012 R2 Datacenter runs as a 64-bit operating system on the server.

\subsection{Results and Interpretation}
\label{sec:evalResults}
In the following we discuss the results of the algorithms on all defined problem instances.
Since we use the ranked score for measuring the quality of the solutions, all hard scores need to be reduced to zero to consider a solution feasible.
In case the hard scores~($H_1,H_2,H_3$) are not down to zero, the algorithm does not find a feasible solution, which we indicate with dashes~(-) in our results table.
Further, the soft scores aim at the matched time windows in the first soft score and the tour length in the second soft score that needs to be minimized. 
The third soft score is only considered by the optimization if the previous scores are reduced to zero.
As this is not the case in any of our evaluation results, we omit this score in our evaluation.
Please keep in mind, that even if pause times and time windows are handled in the Timeline algorithm, no pause time violations can occur as ensured by our algorithm and we only include the time window violations in the score $S_1$.
For better readability, we re-scaled all values by dividing them by 10,000 in our result presentation.
To sum up the evaluation results of all problem instances, we provide Table~\ref{tab:evalsummary}, which states whether time windows are met as well as mean and standard deviations of the tour length~($S_2$) over 30 runs for probabilistic algorithms, that is, the Local Search, GA, and ACO.

\begin{table*}[htb]
\centering
\caption{Summary of the evaluation results for $S_1$ (time windows = TW) and $S_2$ (tour length score) for all algorithms and problem instances. For probabilistic algorithms, the mean and standard deviation values over 30 runs are listed (P = with pause times, PD = with pickup and delivery). The best values are shown in bold.}
\label{tab:evalsummary}
\begin{adjustbox}{max width=\textwidth}
\begin{tabular}{|l|c|r|c|r|c|r|c|r|r|c|r|r|c|r|r|}
\toprule
\textbf{Algorithm} & \multicolumn{2}{c|}{\textbf{Brute Force}} & \multicolumn{2}{c|}{\textbf{Blackbox-I}} & \multicolumn{2}{c|}{\textbf{Blackbox-II}} & \multicolumn{3}{c|}{\textbf{Local Search}} & \multicolumn{3}{c|}{\textbf{GA}} & \multicolumn{3}{c|}{\textbf{ACO}} \\ \midrule
& TW & $S_2$ & TW & $S_2$ & TW & $S_2$ & TW & \multicolumn{2}{c|}{$S_2$} & TW & \multicolumn{2}{c|}{$S_2$} & TW & \multicolumn{2}{c|}{$S_2$} \\ 
 &  &  &  & &  & &  & mean & std &  & mean & std &  & mean & std \\  \midrule
\textbf{TSP-I} & \cmark & \textbf{91.52} & \cmark & 92.55 & \xmark & 90.95 & \cmark & \textbf{91.52} & 0 & \cmark & \textbf{91.52} & 0 & \cmark & 93.87 & 2.06 \\ \midrule
\textbf{TSP-II} & \multicolumn{2}{c|}{-} & \cmark & 161.87 & \xmark & - & \cmark & 157.62 & 5.68 &\cmark & \textbf{156.74} & 0.89 & \xmark & 222.00 & 7.50 \\ \midrule
\textbf{TSP-II-P} & \multicolumn{2}{c|}{-} & \xmark & 161.87 & \xmark & - & 19/30 & 207.41 & 30.50 & 25/30 & 190.53 & 20.26 &\xmark & 217.74 & 7.58\\ \midrule
\textbf{VRP-I} & \multicolumn{2}{c|}{-} & \cmark & 187.14 & \cmark & 185.71 & \cmark & 186.56 & 8.25 & \cmark & \textbf{177.19} & 1.22 & \cmark & 201.81 & 8.85  \\ \midrule
\textbf{VRP-I-P} & \multicolumn{2}{c|}{-} & \xmark & 187.14 & \xmark & 177.82 & \cmark & 194.10 & 5.84 & \cmark & \textbf{179.81} & 0.67 & \multicolumn{3}{c|}{-} \\ \midrule
\textbf{VRP-II} & \multicolumn{2}{c|}{-} & \cmark & 396.21 & \cmark & 373.05 & 28/30 & 299.03 & 60.95 & \cmark & \textbf{292.86} & 14.60 & \multicolumn{3}{c|}{-} \\ \midrule
\textbf{TSP-PD} & \multicolumn{2}{c|}{-} & \multicolumn{2}{c|}{-} & \multicolumn{2}{c|}{-} & 21/30 & 108.50 & 2.00 & 27/30 & \textbf{104.89} & 17.76 & \multicolumn{3}{c|}{-} \\ \midrule
\textbf{VRP-PD} & \multicolumn{2}{c|}{-} & \multicolumn{2}{c|}{-} & \multicolumn{2}{c|}{-} & \cmark & 337.80 & 18.40 & \cmark & \textbf{332.39} & 13.96 & \multicolumn{3}{c|}{-} \\ \bottomrule
\end{tabular}
\end{adjustbox}
\end{table*}
The ticks~(\cmark) indicate, that all time windows are met in all runs of the algorithm, the crosses~(\xmark) show that these are not met.
A value of 19/30 for the time windows shows that in 19 out of 30 runs, all time windows are met.
We only report results of the Brute Force algorithm for the TSP-I problem instance since it already took the algorithm 7~hours and 15~minutes to find a solution for this problem instance.
For larger problem instances, for example the TSP-II problem instance, the algorithm has to assess $33! = 8,6 \cdot 10^{36}$ possible solutions of sequences and, hence, was not able to calculate the optimum solution within feasible time.
Further, we consider a maximum calculation time for all algorithms that is specified by\company{}.
This maximum calculation time is defined to stay within a practically applicable runtime of the algorithms between 60 and 300 seconds.
We decided to set these time limits as we want to be able to react to changes in the orders, vehicles, stops at any point in time and we do not assume that the rVRP is planned once at the beginning of the day, but should be adjustable at any time.
The mean and standard deviation values in the table are calculated using the final score values of the solutions provided after the execution time.
Both Blackbox algorithms are tested for all problem instances except for the pickup and delivery instances since they are not designed to handle pickup and delivery problems.
Additionally, we provide line charts and box plots for all problem instances.
The line charts represent the course of the mean values over 30 repetitions of the $S_2$ score throughout the optimization.
For non-deterministic algorithms (LS, GA, ACO) we further show the standard deviations as error bars.
The box plots represent the final $S_2$ results of the algorithms after the execution time is over. 
To make statements on statistical significance of the results we perform Wilcoxon signed rank tests for the non-deterministic algorithms in all relevant comparisons.
We define the null hypotheses to be that the mean values are drawn from the same distribution and, hence, have no statistically significant difference.
Further, we define the significance level to be $\alpha = 0.05$.
In the following, we first present the results for the TSP instances, then the ones for the VRP instances.

\begin{figure}[htb]
    \centering
    \includegraphics[width=0.99\columnwidth]{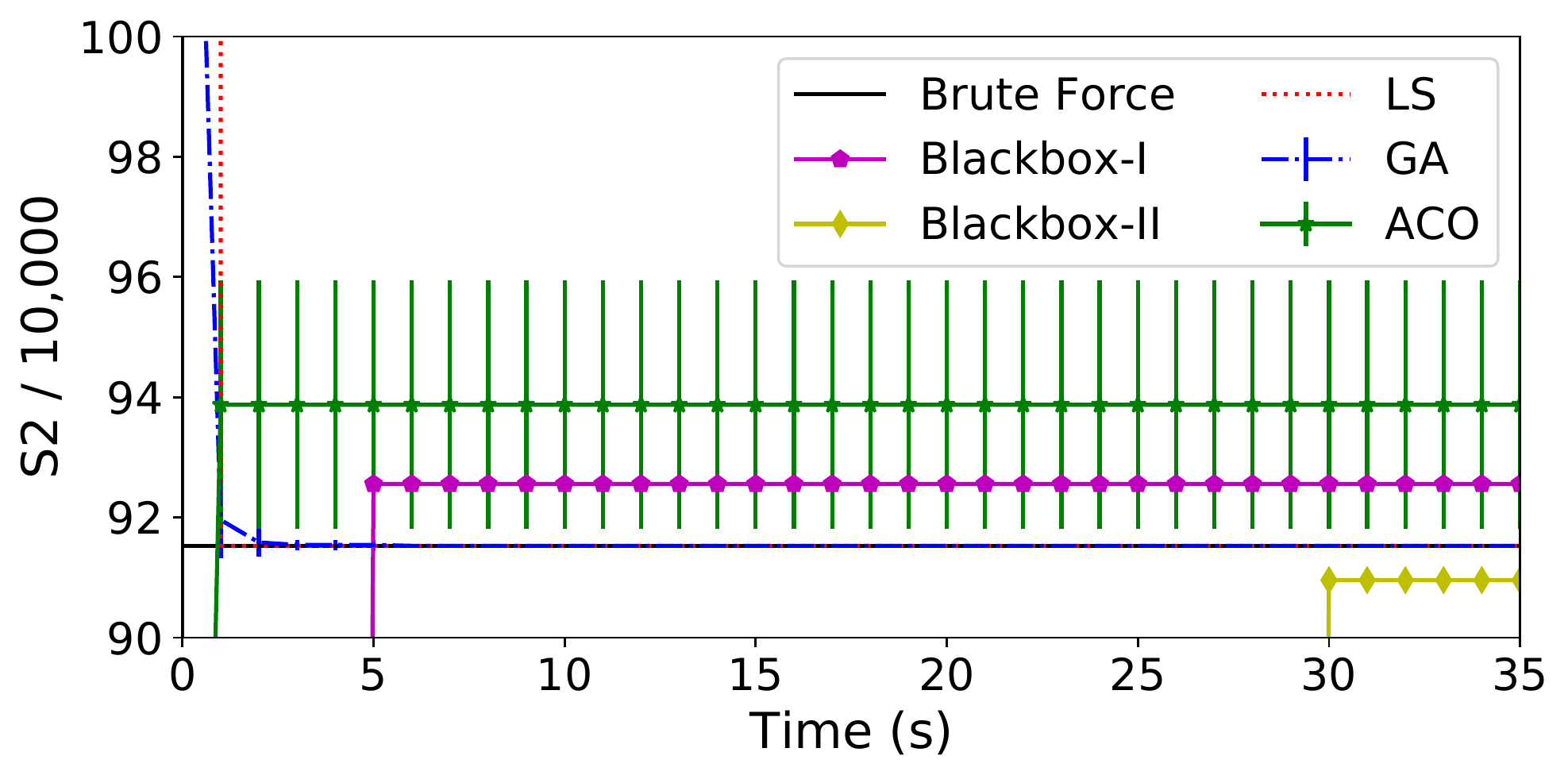}
    \caption{Mean and standard deviations of tour length score ($S_2$) to be minimized for the TSP-I problem instance for all algorithms.}
    \label{fig:tsp1_time}
\end{figure}

\begin{figure}[htb]
    \centering
    \includegraphics[width=0.99\columnwidth]{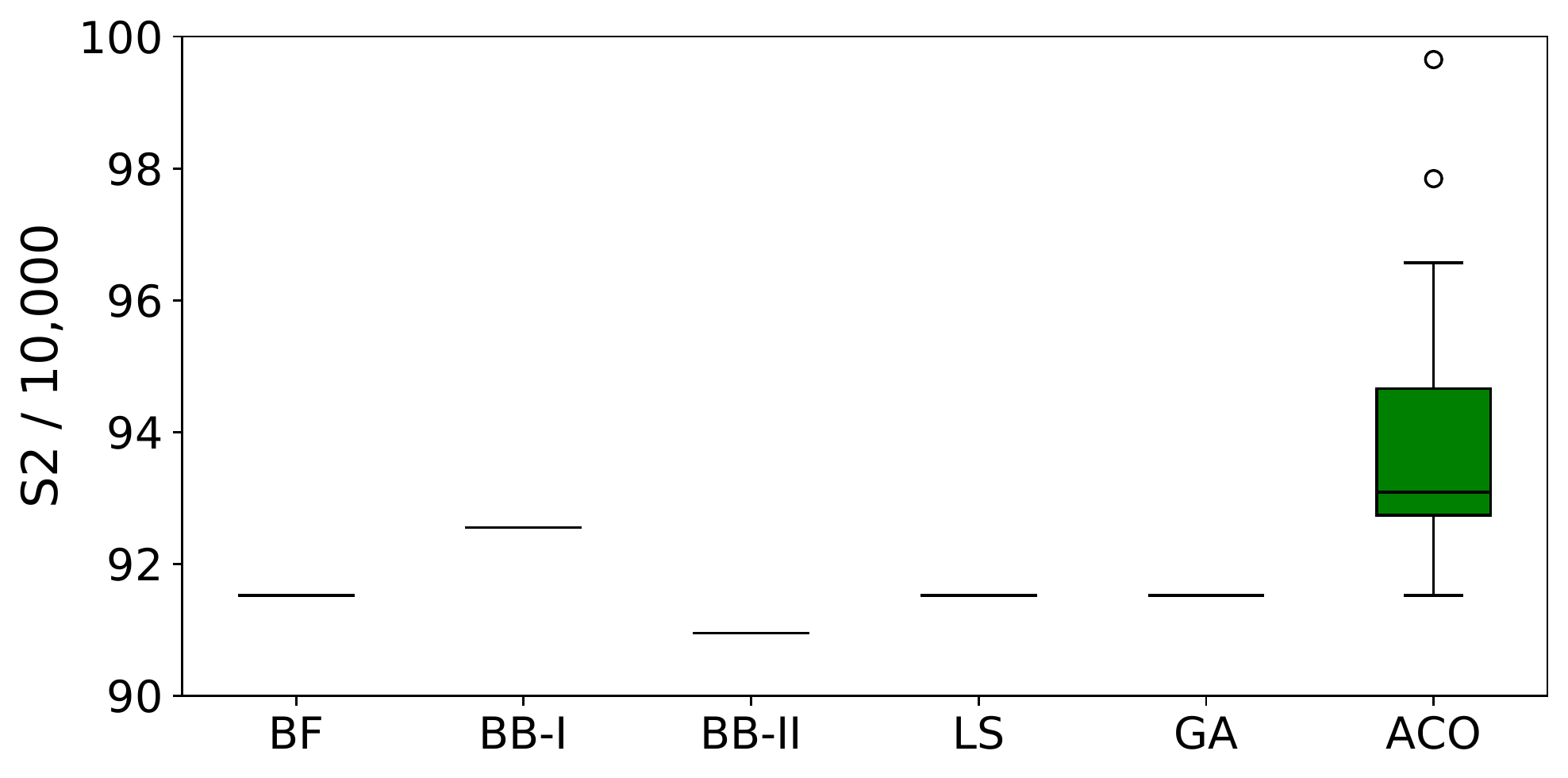}
    \caption{Boxplot of the tour length score ($S_2$) to be minimized for the TSP-I problem instance for all algorithms.}
    \label{fig:tsp1_box}
\end{figure}

The table shows that for the TSP-I problem instance, the LS and GA are able to fulfill all time windows and find the best possible score value (determined by the result of the brute force algorithm).
The Blackbox-II algorithm finds a solution with a reduced score value of around 2.6 score points less but was not able to fit the time windows.
The Blackbox-I and ACO algorithms are able to fit all time windows but only find solutions with higher score value, that is, around 1.00 and 2.35 score points above the optimal score value, respectively.
Figure~\ref{fig:tsp1_time} shows the mean and standard deviation values of the $S_2$ score for all algorithms during the course of optimization.
The x-axis shows execution time in seconds while the y-axis presents the $S_2$ score value divided by 10.000 to achieve better visibility of the values. 
The Brute Force algorithm is depicted as constant line at 91 score points for better comparability with  the other algorithms even if it took more than seven hours to return the result.
Both Blackbox algorithms do not provide the possibility to show the course of optimization but provide a final result after their calculation.
The result of the Blackbox-I is returned after five seconds with a lower value of 90.05 than the optimal one of 91.52 but with broken time windows, while the Blackbox-II algorithm requires 30 seconds calculation time and matches the time windows. 
The LS and GA show very fast convergence towards the optimum solution in all repetitions while the ACO algorithm is not able to achieve the best solution and shows comparably high standard deviations of two score points.
This can also be observed in the box plot in Figure~\ref{fig:tsp1_box}.
Since LS and GA computed the optimal solutions in all repetitions, we do no performed statistical tests on this problem instance.
In summary, LS and GA were able to achieve the best possible solution after only a few seconds and in all runs while the Blackbox algorithms produce worse solutions and the ACO cannot compete with the other algorithms.

\begin{figure}[htb]
    \centering
    \includegraphics[width=0.99\columnwidth]{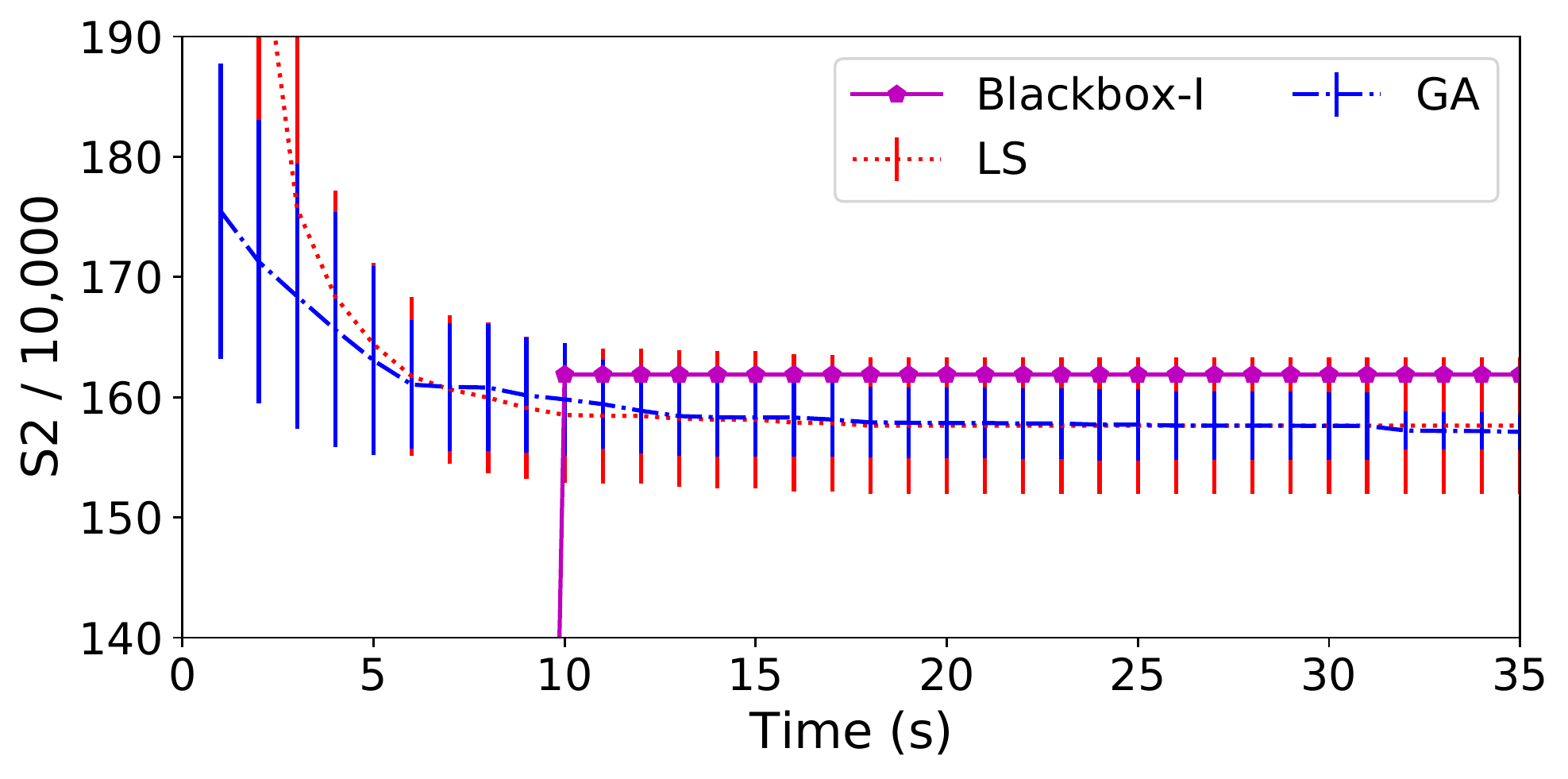}
    \caption{Mean and standard deviations of tour length score ($S_2$) to be minimized for the TSP-II problem instance for all algorithms.}
    \label{fig:tsp2_time}
\end{figure}

\begin{figure}[htb]
    \centering
    \includegraphics[width=0.99\columnwidth]{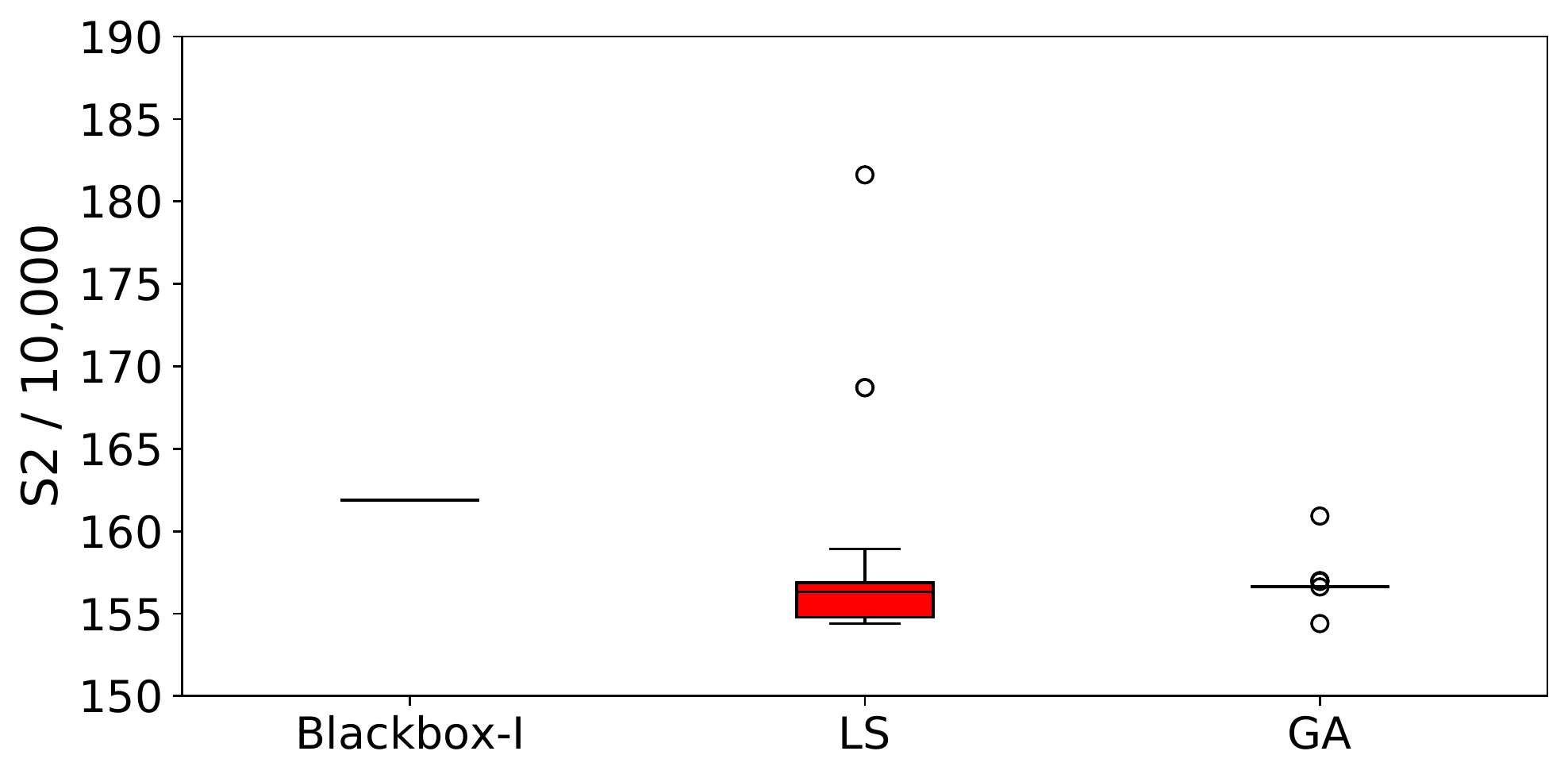}
    \caption{Boxplot of the our length score ($S_2$) to be minimized for the TSP-II problem instance for all algorithms.}
    \label{fig:tsp2_box}
\end{figure}

For the TSP-II instance, the Brute Force algorithm was not able to calculate the optimal solution, while Blackbox-II and ACO were not able to match the time windows. 
The Blackbox-I, LS, and GA find solutions that match all time windows and comparable $S_2$ score values of around 160 score points with GA showing the lowest score with 156.74.
Figure~\ref{fig:tsp2_time} shows the mean and standard deviation values of the $S_2$ score for the algorithms that were able to match all time windows during the course of optimization. 
The Blackbox-I again delivers its solution of 161.87 after it finishes its calculation after ten seconds.
In the mean time, the LS and GA were able to reduce their mean value below the value of Blackbox-I and manage to reduce their standard deviations as well. 
Figure~\ref{fig:tsp2_box} shows the results for the three algorithms as box plot. 
The LS shows the larger box and, hence, a wider variety of solutions.
In contrast, the GA shows stable behavior with a slightly slower mean value with a difference of 0.9 score points compared to the LS.
We performed Wilcoxon signed rank tests to check for statistical significance in the test results between the non-deterministic LS and GA.
We define the $H_0$ hypotheses to be that the mean values are drawn from the same distribution and calculated a p-value of 0.185.
Hence, we were not able to reject our hypotheses with a significance level of $\alpha = 0.05$.
In summary, the LS and GA calculate the best solutions after around ten seconds and perform equally well.

\begin{figure}[htb]
    \centering
    \includegraphics[width=0.99\columnwidth]{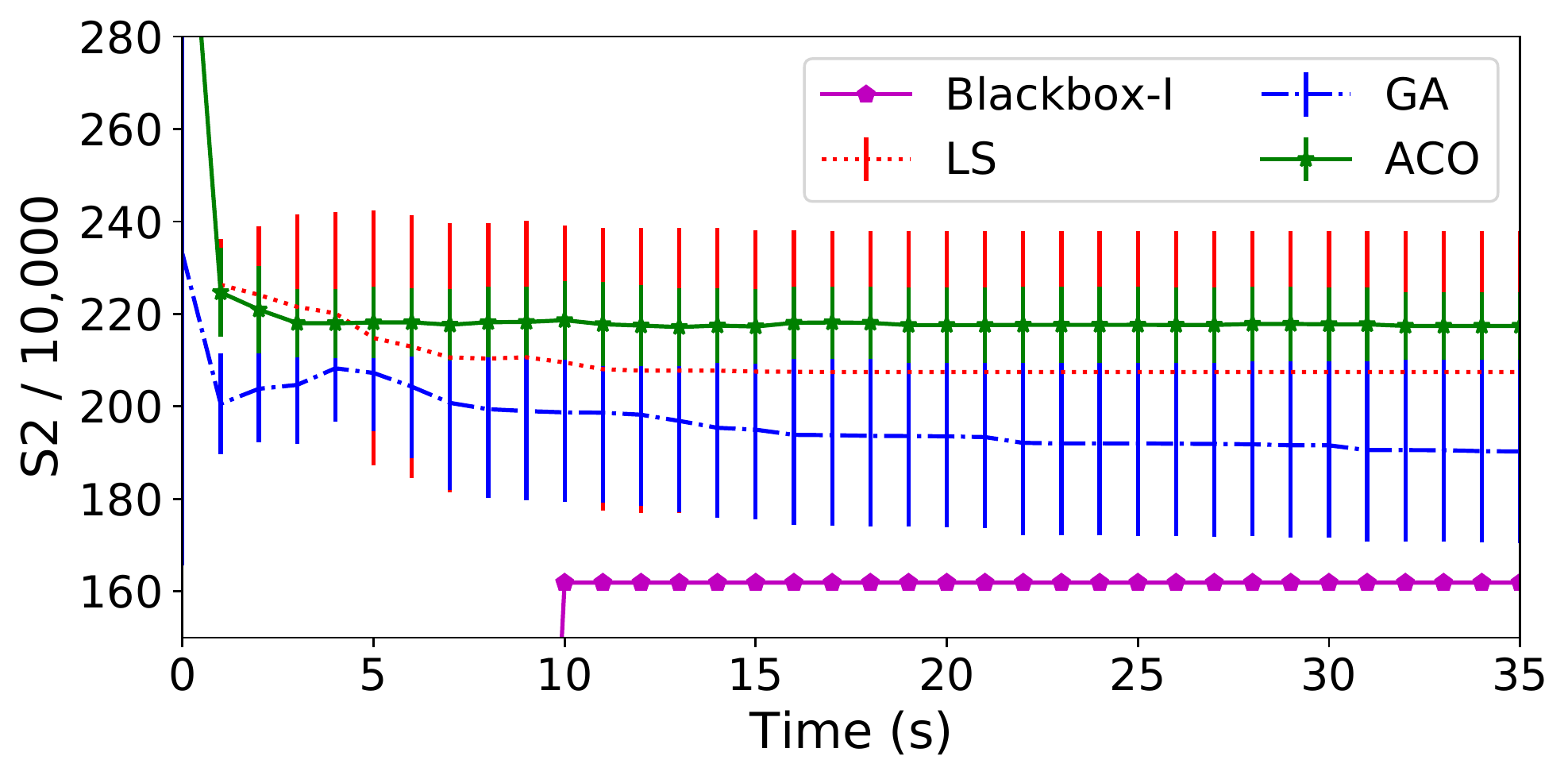}
    \caption{Mean and standard deviations of tour length score ($S_2$) to be minimized for the TSP-II-P problem instance for all algorithms.}
    \label{fig:tsp2p_time}
\end{figure}

\begin{figure}[htb]
    \centering
    \includegraphics[width=0.99\columnwidth]{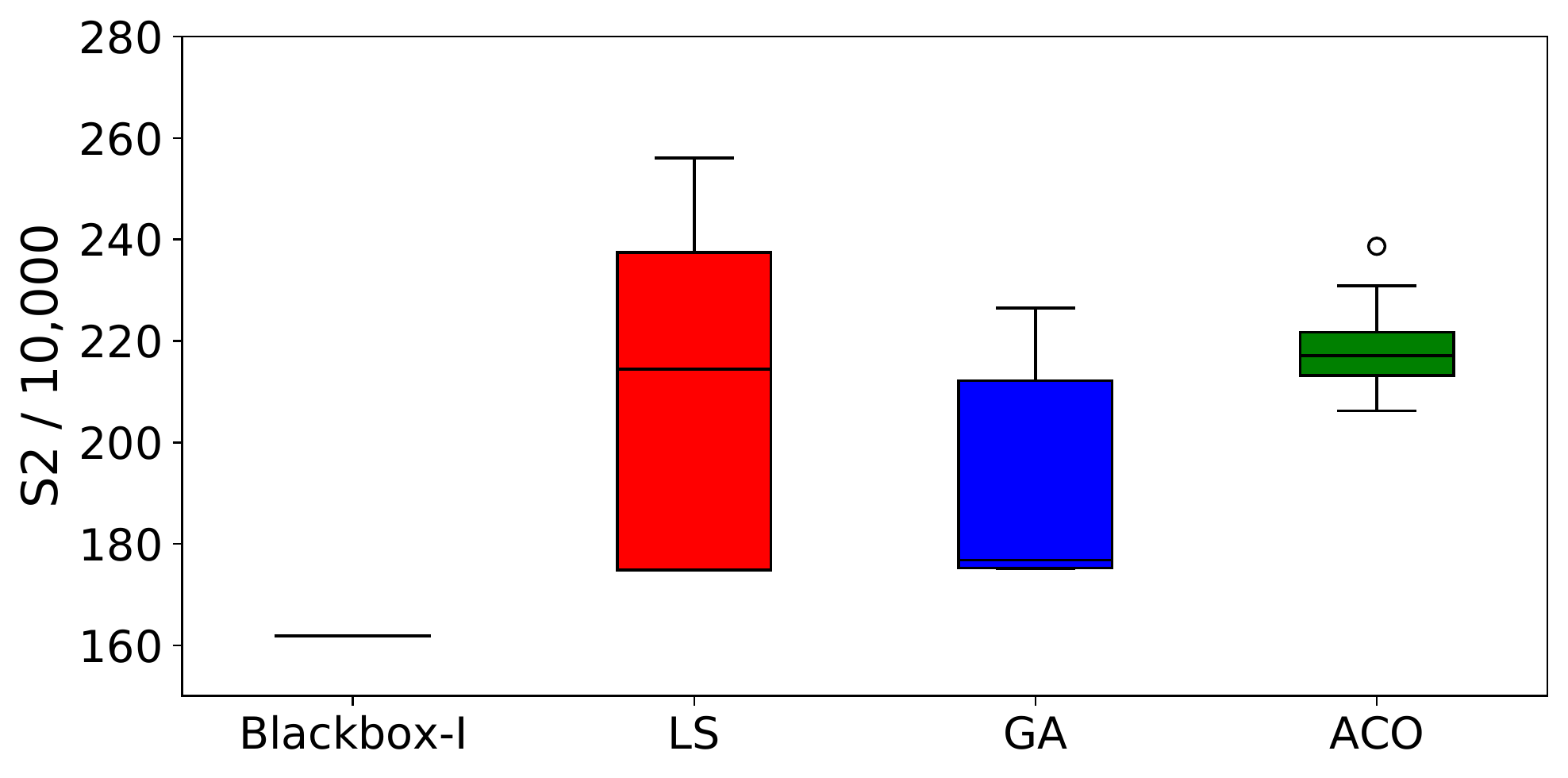}
    \caption{Boxplot of the tour length score ($S_2$) to be minimized for the TSP-II-P problem instance for all algorithms.}
    \label{fig:tsp2p_box}
\end{figure}

When including pause times in the TSP-II-P instance, the Blackbox-I, Blackbox-II, and ACO algorithm were not able to match all time windows in any of the proposed solutions.
Contrary, the LS was able to match time windows in 19 out of 30 solutions and the GA in 25 of 30 solutions.
Additionally, the GA produces results with lower $S_2$ score of 190 score points compared to 207 score points and lower standard deviation~(20 score points for GA and 30 score points for LS).
Again, Figure~\ref{fig:tsp2p_time} provides the mean and standard deviation values of the $S_2$ score for the Blackbox-I, LS, GA, and ACO algorithms. 
Since no algorithm was able to match all time windows in all runs, we present the results of all algorithms and keep the performance regarding the time windows in mind.
The Blackbox-I algorithm returns its solution of 161 score points after ten seconds and has a lower $S_2$ score compared to the other algorithms. 
However, as this solution does not match any time window, we consider it worse than the other algorithms.
The course of optimization of the LS, GA, and ACO show, that the GA already starts with a better value~(230 score points) than both other algorithms~(higher than 280 score points) and continues to decrease the score slightly during runtime. 
The LS is also able to decrease its score but a high standard deviation of 8 score points compared to 1 score point for the GA can be observed while the ACO seems to show no improvement at all.
This can be explained by the fact that the ACO was not able to match the time windows in any run and hence, does not focus on optimizing the $S_2$ score.
The boxplots in Figure~\ref{fig:tsp2p_box} show similar results with a high mean value for the LS and the ACO and a low mean for the GA.
As the ACO was not able to match the time windows in any run, we consider its performance worse than the ones from LS and GA.
We again performed a Wilcoxon signed rank test to compare LS and GA and calculated a p-value of 0.082 and were not able to reject our hypotheses with a significance level of $\alpha = 0.05$.
In summary, all algorithms were not able to find solutions with matching time windows in all repetitions.
However, the LS and GA were able to match time windows in some of the repetitions and are considered best performing in this problem instance.

\begin{figure}[htb]
    \centering
    \includegraphics[width=0.99\columnwidth]{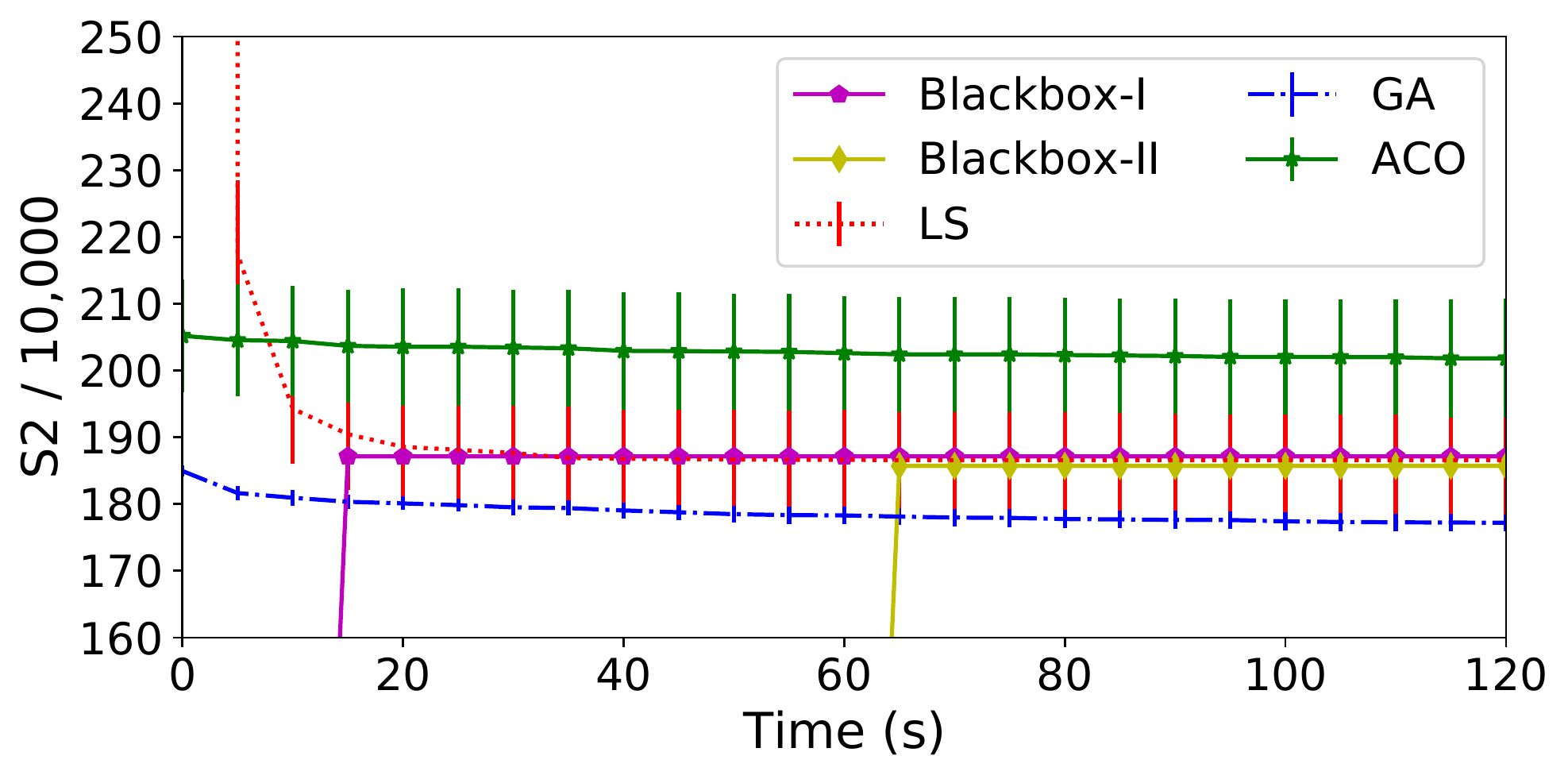}
    \caption{Mean and standard deviations of tour length score ($S_2$) to be minimized for the VRP-I problem instance for all algorithms.}
    \label{fig:vrp1_time}
\end{figure}

\begin{figure}[htb]
    \centering
    \includegraphics[width=0.99\columnwidth]{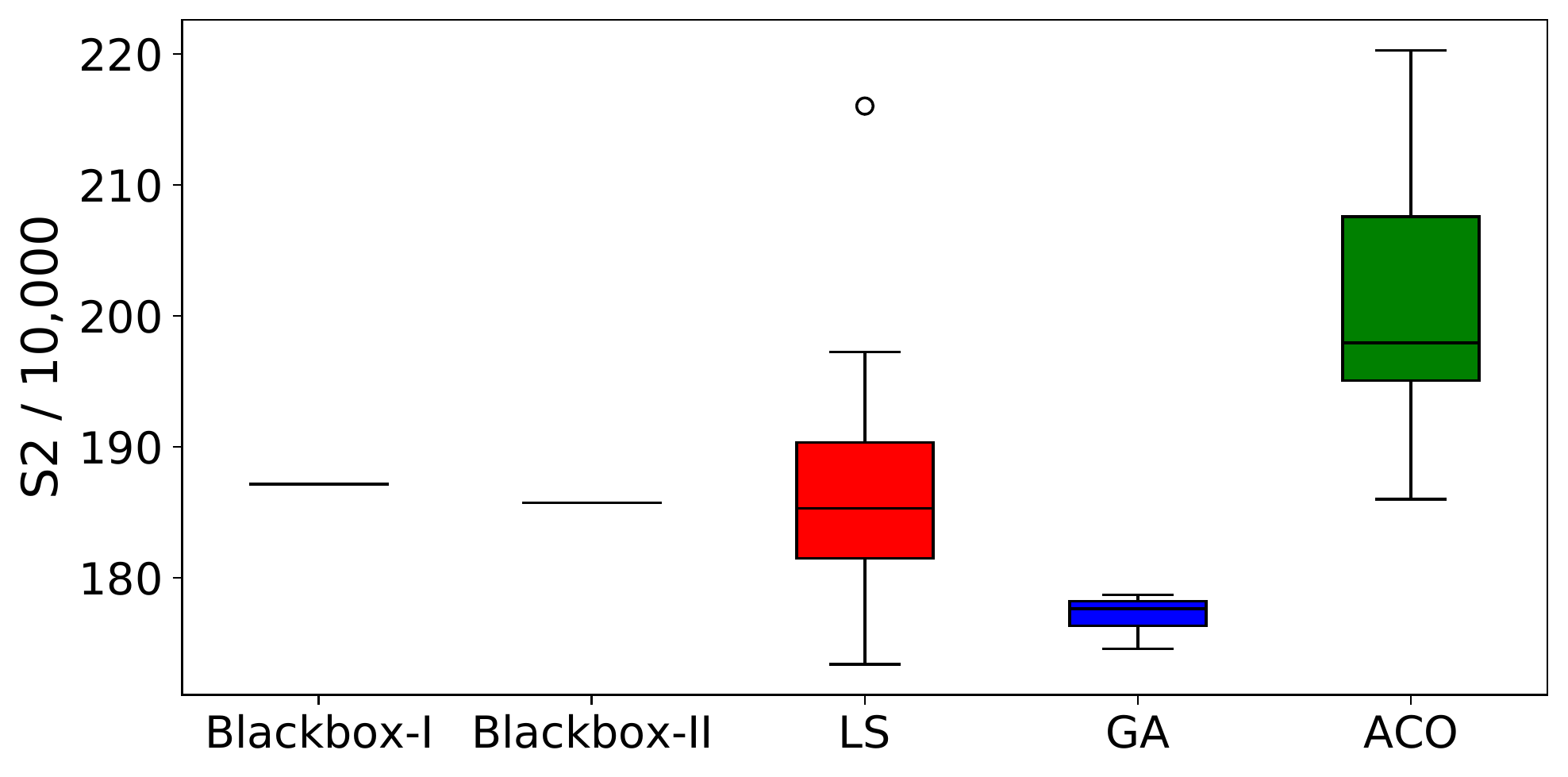}
    \caption{Boxplot of the tour length score ($S_2$) to be minimized for the VRP-I problem instance for all algorithms.}
    \label{fig:vrp1_box}
\end{figure}

The results using the first VRP problem instance~(VRP-I) show, that all tested algorithms are able to match the time windows.
While the ACO provides solutions with a high $S_2$ score of around 200 score points, the scores of the other algorithms are comparably low at around 186 score points with the GA showing the lowest value of 177 score points.
The line chart in Figure~\ref{fig:vrp1_time} shows the course of optimization for all algorithms.
Blackbox-I delivers its result of 187 score points after around 18 seconds while the Blackbox-II algorithm requires 65 seconds calculation time with a score value of 186 score points. 
Both algorithms deliver results with higher $S_2$ score value compared to the GA with a score of 177. 
The GA already starts with a good initialized value of around 185 and continues reducing the $S_2$ score over time with small standard deviations of around 1 score point.
The LS algorithm starts with a high mean value~(larger than 250 score points), but reduces the score to the level of the Blackbox algorithms in the first 20 seconds.
The ACO algorithm shows higher score values of around 208 score points compared to the other algorithms but at least slightly reduces the score over time. 
A similar result can be seen in Figure~\ref{fig:vrp1_box} where the GA shows the lowest values and the smallest box indicating a very stable low score value.
The mean of the LS is around the values for the Blackbox algorithms but has a larger box and hence, shows a larger diversity in the results.
Finally, the ACO has a higher mean value and a larger variety in the results.
Using three Wilcoxon signed rank tests between LS, GA, and ACO, we are able to reject the hypotheses with p-values of 0.001 and a significance level of $\alpha = 0.05$, which means that the mean values are drawn from different distributions and hence the difference is statistically significant between all algorithms.
In summary, the GA shows significant improvements over the LS and ACO algorithms in this problem instance.

\begin{figure}[htb]
    \centering
    \includegraphics[width=0.99\columnwidth]{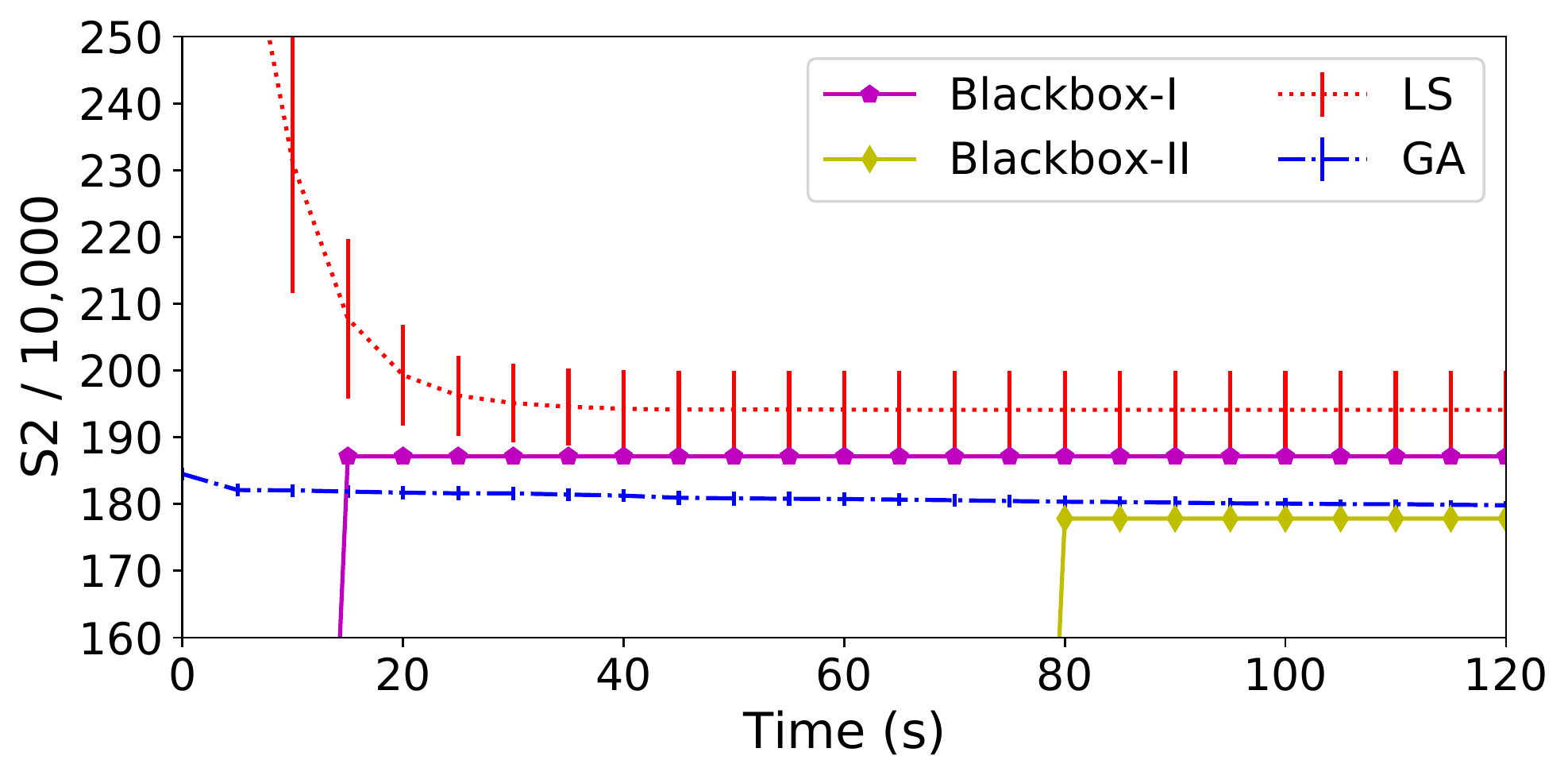}
    \caption{Mean and standard deviations of tour length score ($S_2$) to be minimized for the VRP-I-P problem instance for all algorithms.}
    \label{fig:vrp1p_time}
\end{figure}

\begin{figure}[htb]
    \centering
    \includegraphics[width=0.99\columnwidth]{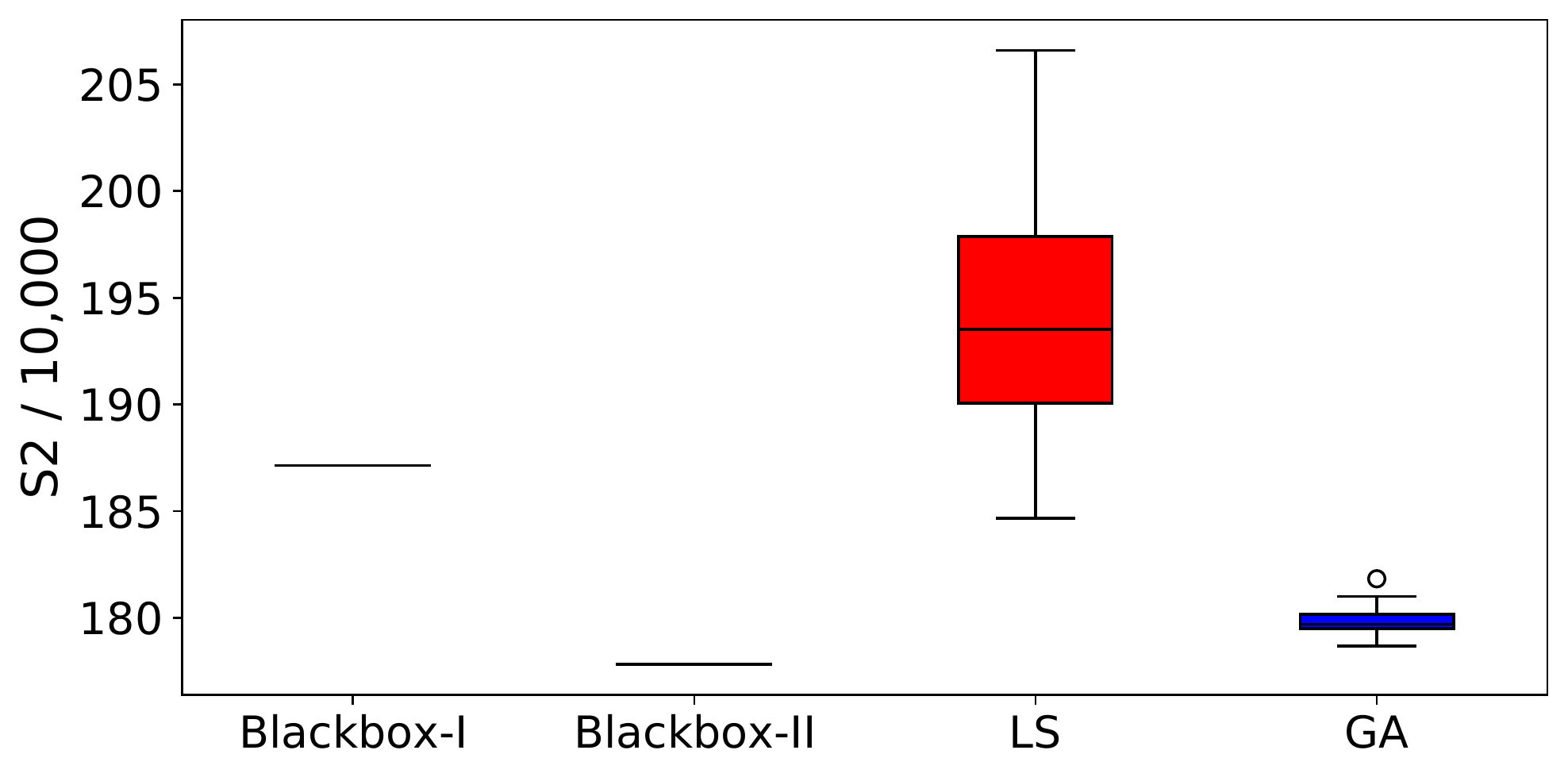}
    \caption{Boxplot of the tour length score ($S_2$) to be minimized for the VRP-I-P problem instance for all algorithms.}
    \label{fig:vrp1p_box}
\end{figure}

For the VRP-I-P problem instance with pause times, both Blackbox algorithms are unable to match the time windows.
On the contrary, the LS and GA solutions match all time windows with GA having the lowest mean of 180 score points and standard deviation of 0.7 score points.
The course of the optimization is depicted in Figure~\ref{fig:vrp1p_time}. 
The Blackbox-I delivers its result of 187 score points after around 18 seconds while Blackbox-II algorithm requires 80 seconds calculation time with a final score of 177. 
Again, the GA starts with an already good solution of around 185 score points and further reduces the score value throughout the calculation time while the LS starts with a very high score larger than 250 score points.
The LS is able to reduce its score within the first 30 seconds but still has a difference of around 15 score points to the GA. 
The boxplot in Figure~\ref{fig:vrp1p_box} supports this finding, as the GA has low score value and produces very stable results with only few variations, while the LS shows worse results and high variability in the solution quality. 
In addition, a Wilcoxon signed rank test calculates a p-value of 0.001 and we are able to reject the hypotheses with a significance level of $\alpha = 0.05$, and hence the mean values are different.
In summary, the GA shows significant improvements over the LS and shows the most stable solution quality.

\begin{figure}[htb]
    \centering
    \includegraphics[width=0.99\columnwidth]{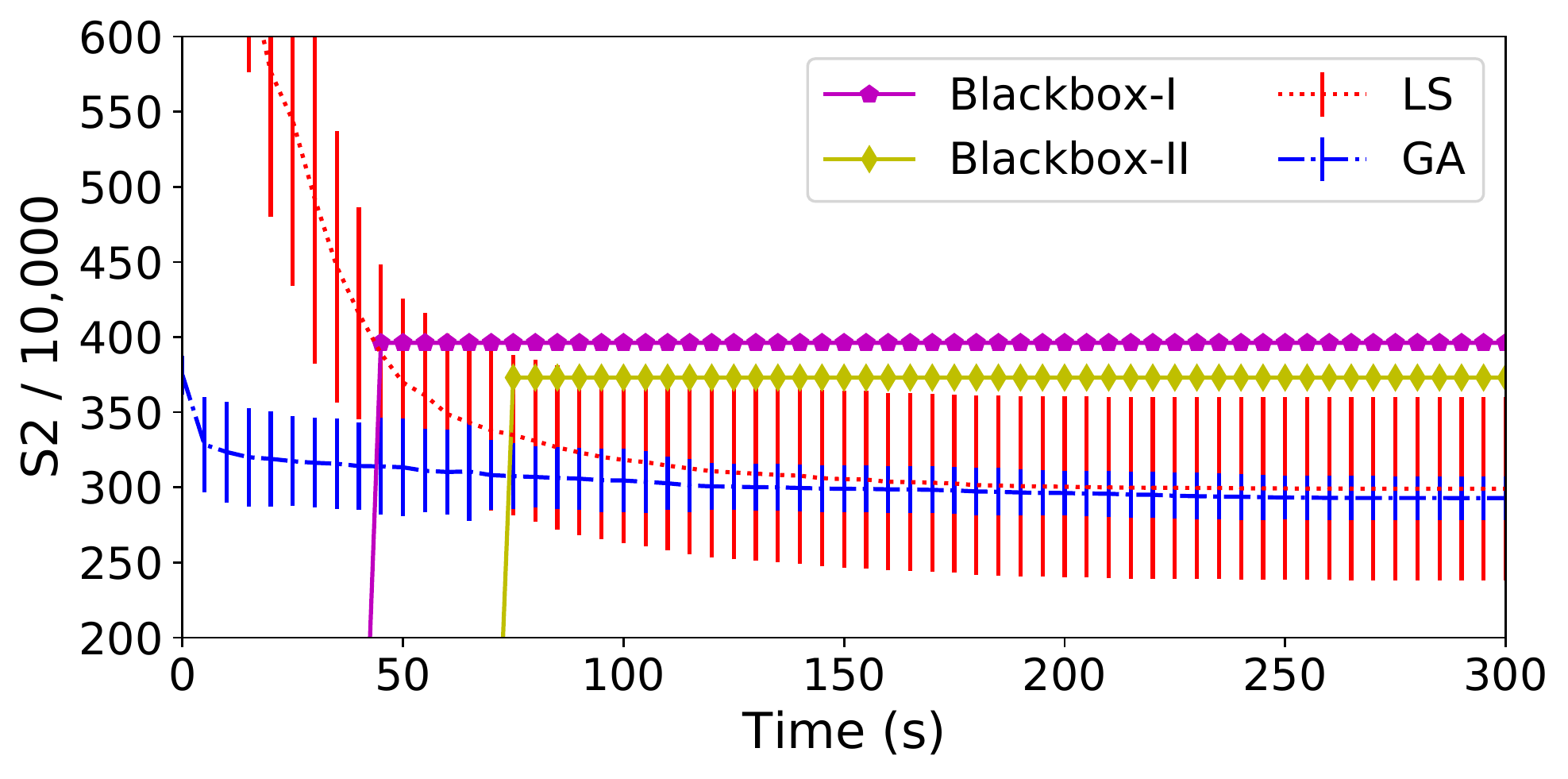}
    \caption{Mean and standard deviations of tour length score ($S_2$) to be minimized for the VRP-II problem instance for all algorithms.}
    \label{fig:vrp2_time}
\end{figure}

In the second VRP problem instance~(VRP-II), both Blackbox algorithms are able to match all time windows, while LS only matches time windows in 28 of 30 runs.
The solutions of the GA match all time windows in all runs, and hence, are considered better than the solutions of the LS.
Figure~\ref{fig:vrp2_time} shows the course of optimization during calculation time. 

\begin{figure}[htb]
    \centering
    \includegraphics[width=0.99\columnwidth]{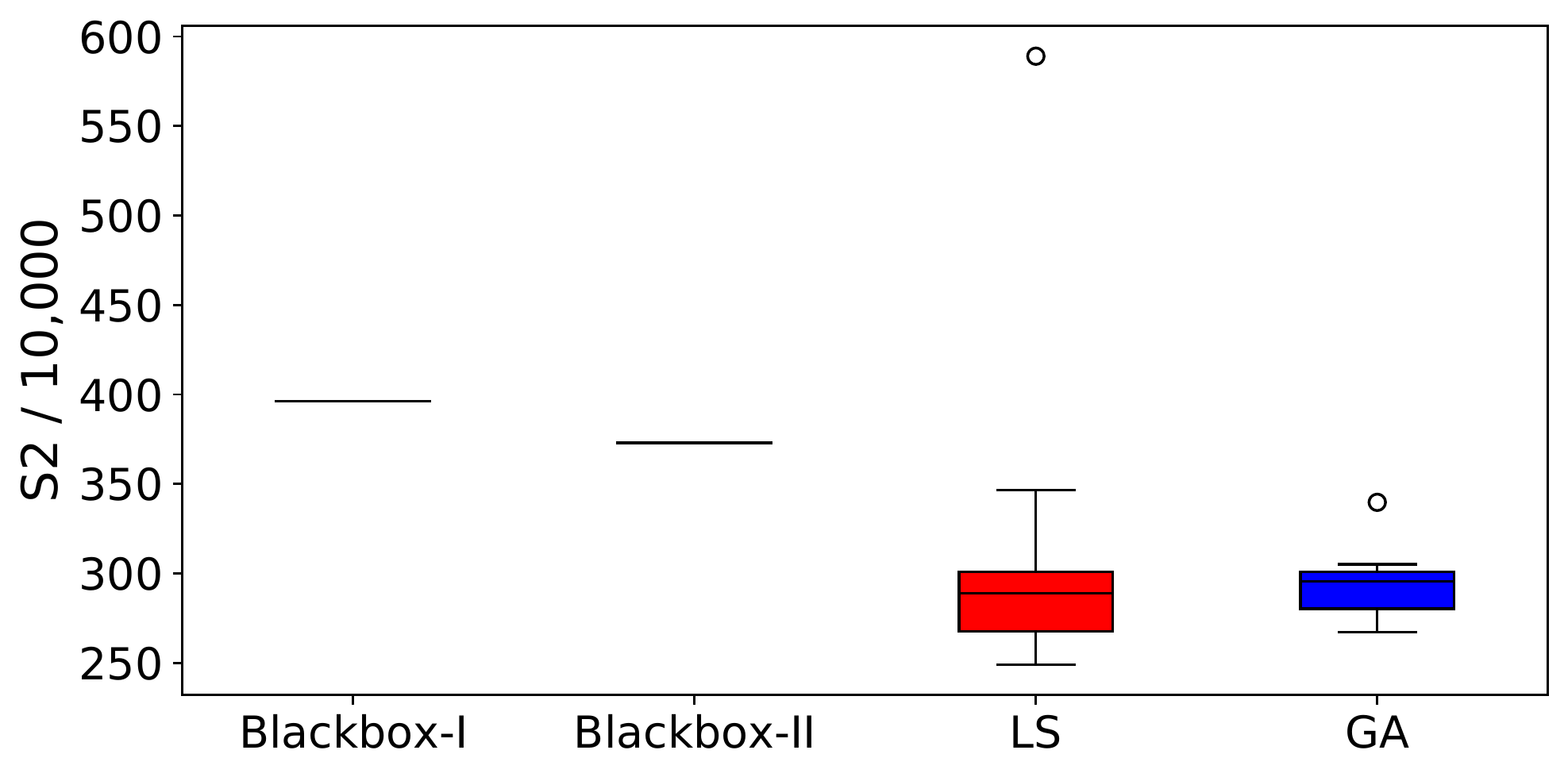}
    \caption{Boxplot of the tour length score ($S_2$) to be minimized for the VRP-II problem instance for all algorithms.}
    \label{fig:vrp2_box}
\end{figure}

The Blackbox-I algorithm returns its result of 396 score points after around 45 seconds, while the Blackbox-II algorithm delivers solutions after around 75 seconds with a score of 373. 
The GA starts with a solution quality in the area of the Blackbox algorithms and further reduces the score value to 293 and its standard deviation to 15 over time.
In contrast, the LS starts with a high score above 600 score points and reduces its solution to the level of the GA at around 100 seconds but shows a higher standard deviation of 61 score points. 
The box plots in Figure~\ref{fig:vrp2_box} shows that the solution quality and the variety of the quality of the final results of both algorithms is comparably good. 
The Wilcoxon signed rank test calculates a p-value of 0.478 and we are not able to reject our hypotheses. 
In summary, the LS and GA algorithms outperform both Blackbox algorithms with regards to the second score and perform comparably good.
However, the LS does not match the time windows in all runs and hence, is considered worse than the GA. 

\begin{figure}[htb]
    \centering
    \includegraphics[width=0.99\columnwidth]{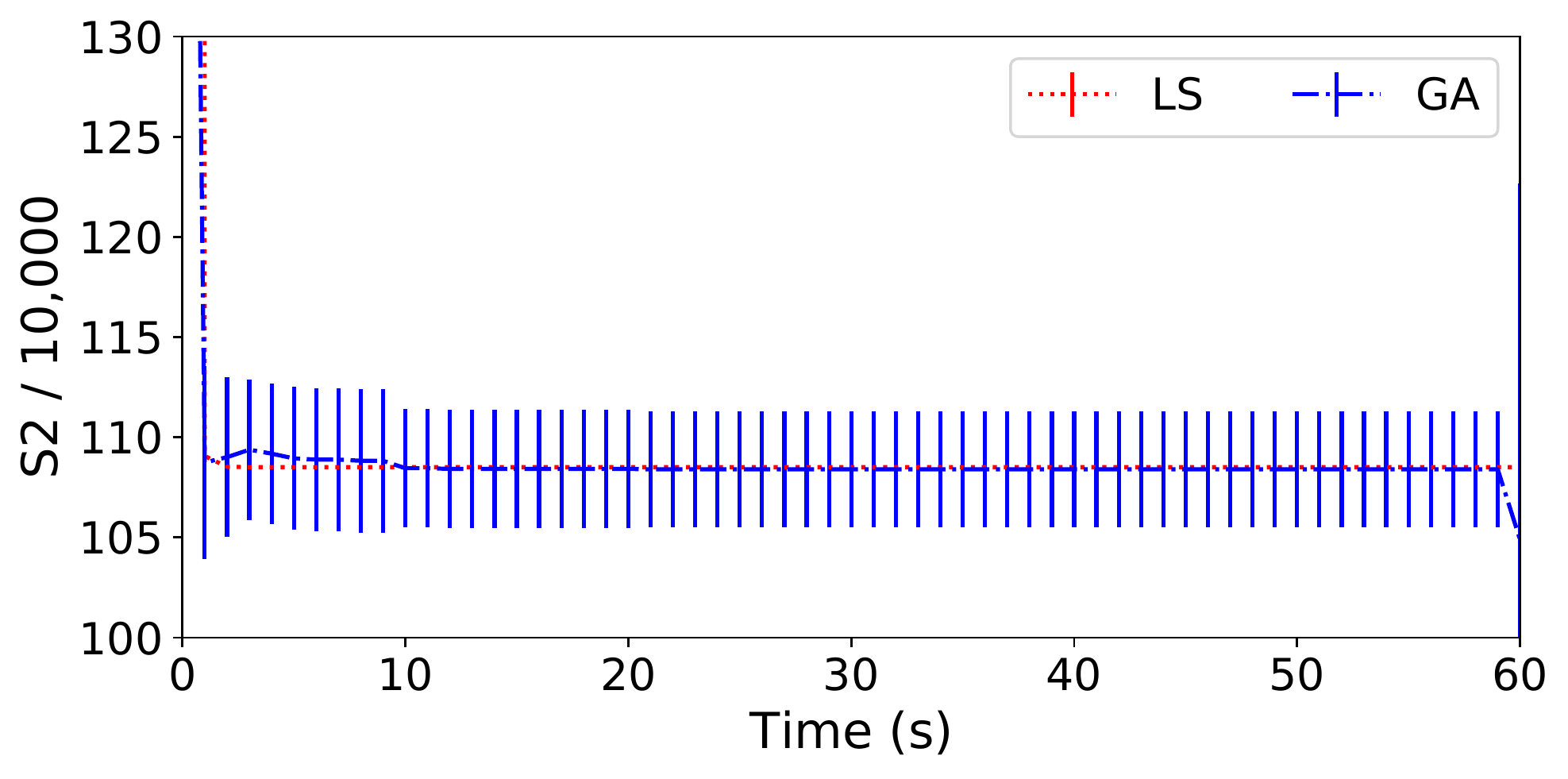}
    \caption{Mean and standard deviations of tour length score ($S_2$) to be minimized for the TSP-PD problem instance for all algorithms.}
    \label{fig:tsppd_time}
\end{figure}

\begin{figure}[htb]
    \centering
    \includegraphics[width=0.99\columnwidth]{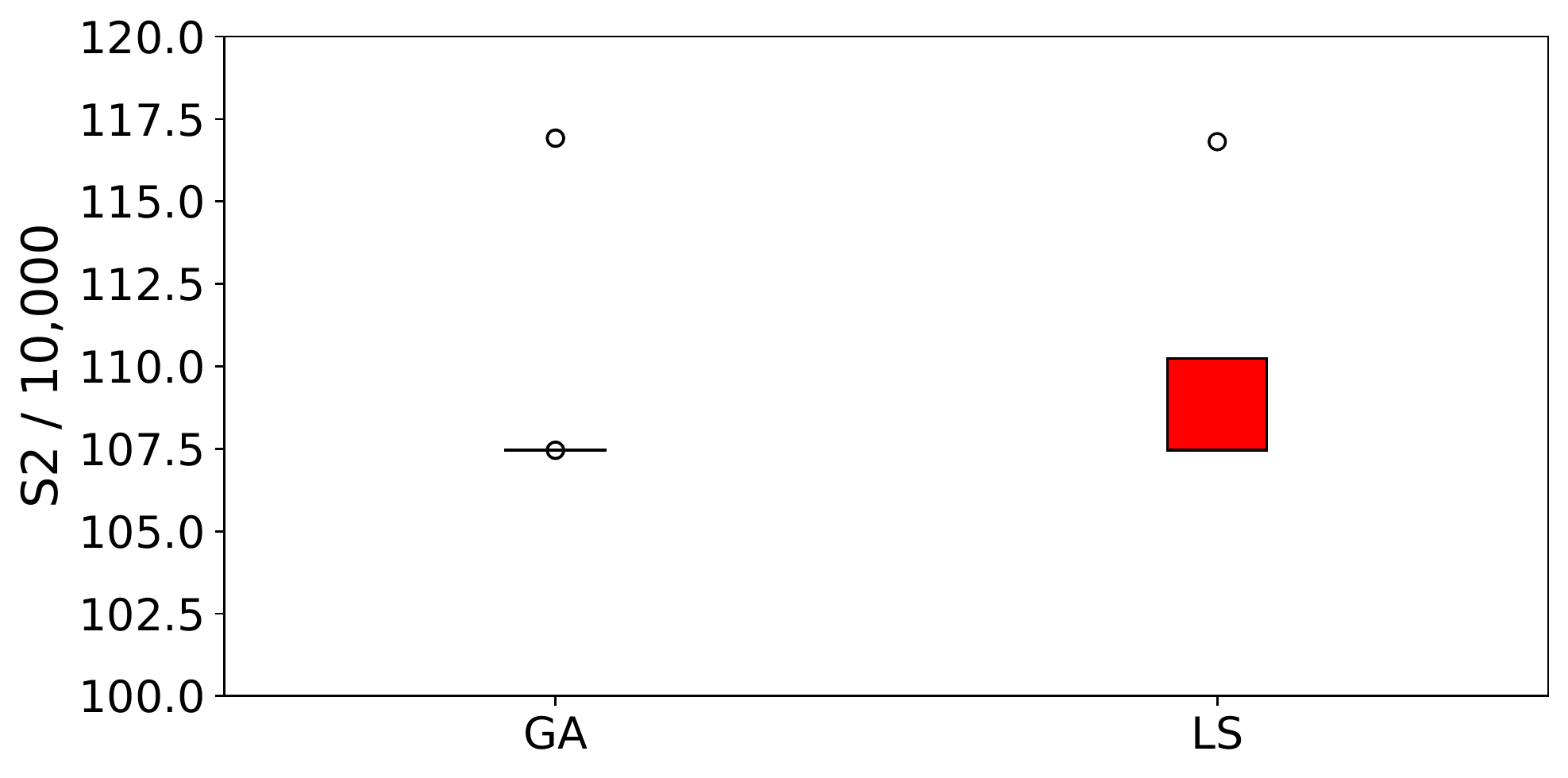}
    \caption{Boxplot of the tour length score ($S_2$) to be minimized for the TSP-PD problem instance for all algorithms}
    \label{fig:tsppd_box}
\end{figure}

For both PD problem instances, we compare the LS and the GA since the Blackbox algorithms cannot handle PD problems. 
In the TSP-PD problem instance, LS matches time windows in 21 out of 30 runs, and the GA in 27 out of 30 runs.
Hence, the GA can be considered more stable than the LS as the probability to receive solutions with matching time windows is higher. 
Figure~\ref{fig:tsppd_time} shows the optimization result during the runtime of the algorithms.
Both algorithms start with high score values above 130 and reduce the score in the first two to three seconds to values around 108 for the LS and 105 for the GA.
The GA shows larger standard deviations of around 18 score points compared to the LS with a standard deviation of two.
Figure~\ref{fig:tsppd_box} presents box plots of the final results to compare GA and LS.
The box plot of the GA is very small besides one outlier and the LS box plot is larger spreading from 1,080,000 to 1,100,000 score points.
The Wilcoxon signed rank test was not able to reject the hypotheses with a p-value of 0.145.
In summary, both algorithms perform comparably good in this problem instance as both do not match all time windows and deliver nearly the same quality in the $S_2$ score.

\begin{figure}[htb]
    \centering
    \includegraphics[width=0.99\columnwidth]{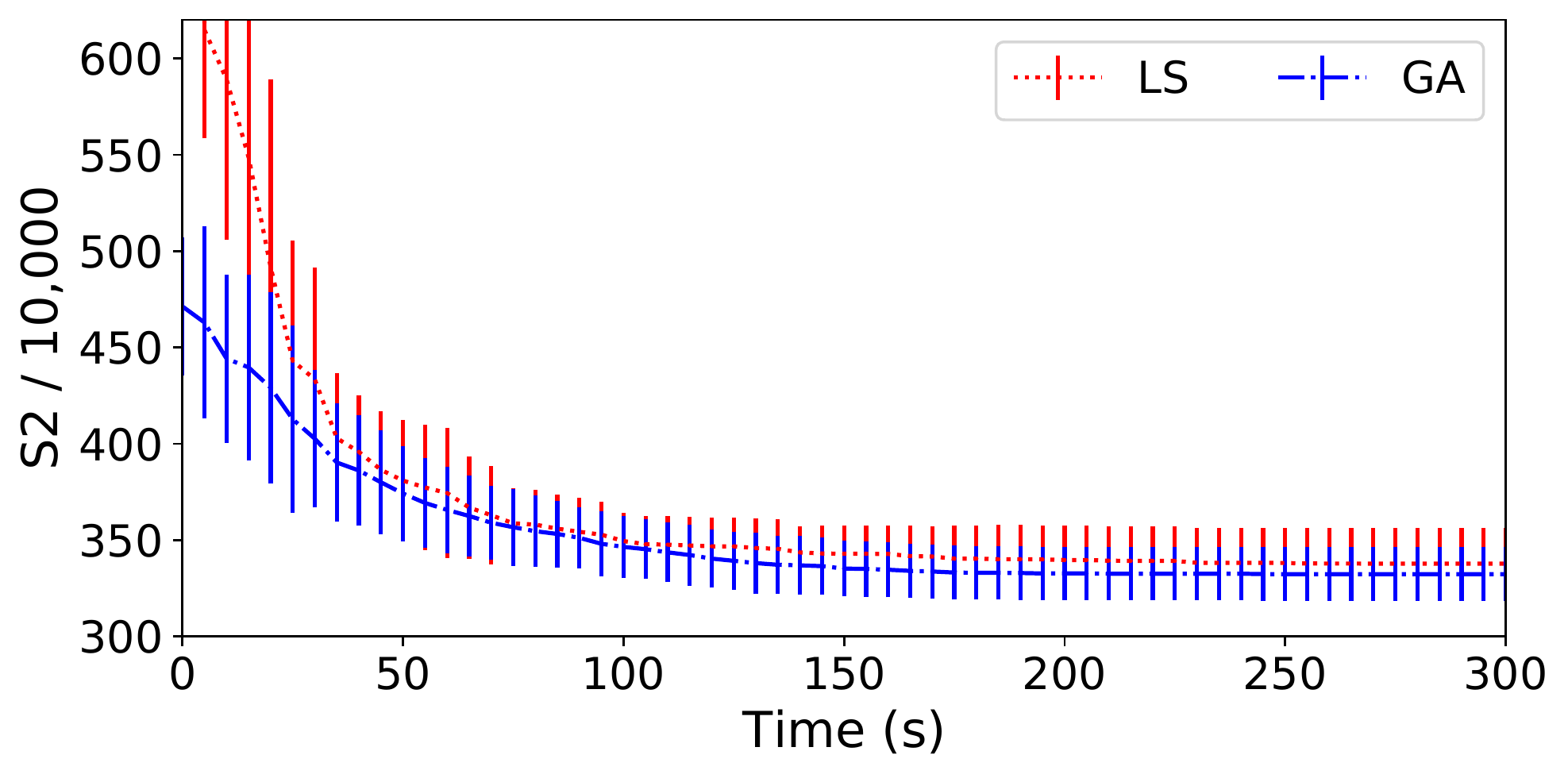}
    \caption{Mean and standard deviations of tour length score ($S_2$) to be minimized for the VRP-PD problem instance for all algorithms.}
    \label{fig:vrppd_time}
\end{figure}

\begin{figure}[htb]
    \centering
    \includegraphics[width=0.99\columnwidth]{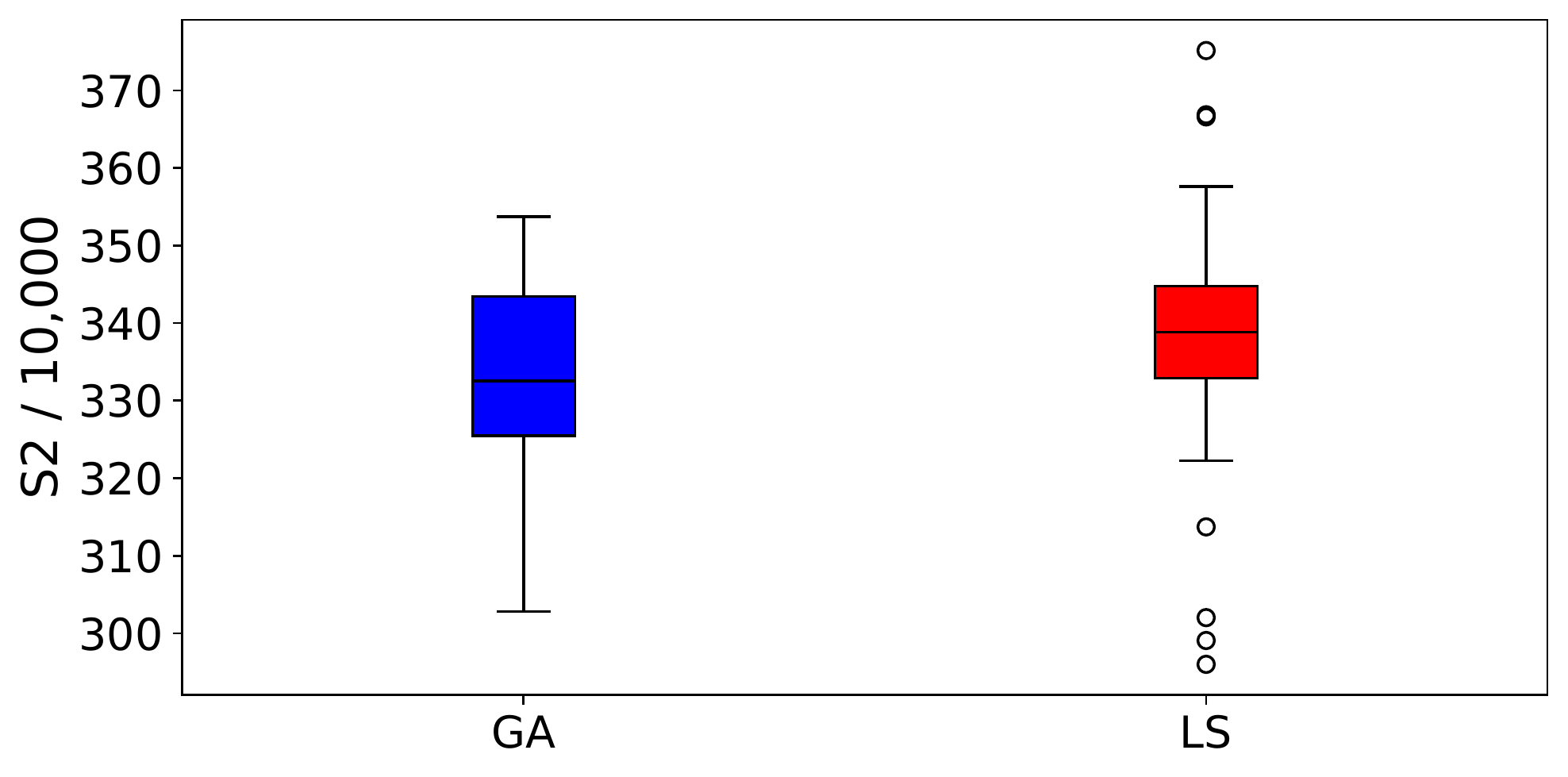}
    \caption{Boxplot of the tour length score ($S_2$) to be minimized for the VRP-PD problem instance for all algorithms.}
    \label{fig:vrppd_box}
\end{figure}

Finally, LS and GA are able to match all time windows in the VRP-PD problem instance.
Further, the GA produces solutions with a lower mean $S_2$ score value of around 332 score points compared to the LS with a value of 338. 
In Figure~\ref{fig:vrppd_time} the GA start with a high value of 450.00 and the LS with a value of 600.00 score points.
But both algorithms decrease the score in the first 100 seconds to around 350.00 score points.
Still, the GA maintains its advance and the mean stays below the mean of the LS.
The standard deviation of both algorithms are similar around 14 to 18 score points.
The box plots in Figure~\ref{fig:vrppd_box} show that the mean values are also very similar.
The box and whiskers of the GA span a wider range, while the LS has a smaller box but some more outliers.
In line, the Wilcoxon signed rank test was not able to reject the hypotheses with a p-value of 0.329.
In summary, again both algorithms perform equally good with a slight advantage of around 50,000 score points for the GA. 

Our evaluation results support the following key findings of our paper.
First, our approach integrates all constraints and requirements given by the real-world application such as setup times, a heterogeneous fleet, multiple depots, P/D, stop options, and return permissions~(Table~\ref{tab:evalsummary}).  
Compared to the existing Blackbox algorithms of\company{}, we additionally support setup times, a heterogeneous fleet, multiple depots, P/D, stop options, and return permissions. 
Second, on the smallest problem instance (TSP-I), the TSP-stage of our GA computes the best possible score confirmed by the Brute Force algorithm~(Figure~\ref{fig:tsp1_time}).
We were not able to prove this behavior for larger problem instances since the brute force computation was not feasible. 
Third, when compared to the existing algorithms Blackbox-I and Blackbox-II, we reduced the time to result from 30 seconds to two seconds for the TSP instance, and from 60 seconds to around five seconds for the VRP instance, which is a computation time reduction of 90\%~(Figures~\ref{fig:tsp1_time} and~\ref{fig:vrp1_time}).
This provides\company{} the possibility to react to unforeseen situations and adapt the tours spontaneously.
Finally, in the largest VRP-PD problem instance our GA algorithm returns meaningful results right from the start contrary to the LS algorithm~(Figure~\ref{fig:vrppd_time}). 
The achieved score of the results from both algorithms shows that they perform on the same level while our GA produces more stable results, indicated by the lower standard deviation~(Figure~\ref{fig:vrppd_box}).
Regarding the GA, the performance could make the impression that the initialization phase of the algorithm already provides good enough solutions and the optimization steps are not able to improve the score. 
However, when looking at all line plots showing the course of the optimization it can be seen, that the score value of the GA can be improved over time. 
Still, the good initialization enables to further reduce the required calculation time and hence, provides the possibility to adapt to changes in a short amount of time. 
In summary, we highlight the following main results of our paper:
\begin{itemize}
    \item Integration of all real-world constraints motivated by and defined with\company{}.
    \item Time-to-result reduction from 30 to two seconds, and from 60 to five seconds for the TSP and VRP instances, respectively, which has enormous practical use.
    \item Same level performance of LS and GA on the largest VRP test instance, while our proposed GA shows the most stable performance.
\end{itemize}

\subsection{Practical Implications}
\label{sec:implications}
\major{
The results as presented in the previous section have several implications for practitioners, as also confirmed by our partner from\company{}.
Next, we will shortly discuss them.
}

\major{
\textbf{Real-world compatible Solutions through the Integration of Constraints.}
In contrast to many approaches in related work~(cf.~Section~\ref{sec:relwork}), we focus on integrating real-world constraints.
We will illustrate the benefits of integrating such constraints on two short examples.
One relevant constraint is the time a driver is allowed to drive.
Obviously, this has a huge impact on the planning procedure. 
Having not included those constraints, this might lead to delays in the planned route due to breaks specified by law. 
In the worst case, a driver is not able to finish the route because the allowed driving time is achieved.
Another constraint are time windows.
Those windows specify the time when it is possible at a customer location to unload the truck.
Without consideration, it might be possible that a driver arrives at a time when unloading is not possible, e.g., because the specific shop is not opened yet.
Our approach does not only take both constraints (besides other, further relevant constraints) into account, but also optimizes them, e.g., if the best possible option would require some waiting time at some location due to time windows, our approach would try to harmonize this with the breaks of drivers.
This is really important to generate plans that are suitable for the practice.
State-of-the-art today is that companies have to adjust the calculated plans to optimize them w.r.t. those constraints; our approach provides a significantly higher level of automation.
}

\major{
\textbf{Preference-optimized Solutions through Multi-objectiveness.}
The used algorithms all support multi-objectiveness. 
Consequently, those can be used to identify Pareto-optimal solutions that are able to optimize several objectives such as time, distance, or invested resources (in terms of drivers, trucks, fuel, etc.) alike.
In literature, one can identify several works that provide a multi-objective approach (e.g.,~\cite{dutta2020multi,barma2021multi,mukherjee2021multi}).
However, none of them also integrates such real-world constraints as mentioned before, which highly complicates the finding of solutions.
Further, to achieve the multi-objectiveness, either metrics as Hypervolume~\cite{zhang2014empirical} that are able to balance the different objectives for reaching a single score (however, with the disadvantage that the balancing process is fixed) or introducing weights for the different objective dimensions is necessary.
Our approach is flexible enough to support dynamic changes of the weights for testing several combinations while being in control of the balance between the objectives. 
This is achieved through the application of nature-inspired algorithms; but also due to the fact that the computational time is rather low.
}

\major{
\textbf{Dynamic Planning through Runtime Performance.}
Our approach was able to reduce the Time-to-result from 30 to 2 seconds and from 60 to 5 seconds for the TSP and VRP instances, respectively.
This sounds only marginal; but taking into account that we have a multi-objective approach with flexible weights for the objectives and that a planner of logistic operations might be interested to test several ratios for balancing the objectives (or the integration of different constraints), a reduction of the runtime for the planning algorithms by a factor of 15 and 20 for the TSP and VRP instances, respectively, this has a huge implication for daily work.
Especially as our scenarios under consideration definitely have the size of real-world scenarios and used real-world data rather than artificially created, potentially biased, data, the short runtime are remarkable.
As we are using heuristic-based approaches, one could argue that the short runtime might lead to reduced quality in the solutions. 
However, we have shown that LS and GA achieve the same level of performance on the largest VRP test instance, while our proposed GA shows the most stable performance.
}

\subsection{Threats to Validity}
\label{sec:threats}
We identified the following threats to validity for our approach.
In this paper, we focus on nature-inspired algorithms~(ACO and GA) for tackling the rVRP and compared them to a Brute-Force, two Blackbox algorithms implemented by\company{}, and local search. 
Those algorithms provide heuristic solutions, which provide fast results, however, require multiple runs to receive reliable results.
Further, we did not evaluate other common algorithms used for these kinds of problems as those often require manual implementation effort to adjust them for the particular rVRP problem as discussed in the beginning of this section.
Therefore, we decided to compare our algorithms to an existing implementation of Local Search inside OptaPlanner.
In the future, we plan to use further algorithms for multi-objective optimization such as NSGA-II, particle swarm, or Branch-and-Bound algorithms. 
Additionally, our results are limited to the defined problem instances and we plan to also evaluate even larger VRP instances in cooperation with\company{} in the future. 
Finally, our analysis of related work showed that existing approaches simplify the problem by using assumptions or neglecting specific aspects. 
One could argue that we over-complicated the problem as so far it has been enough for the industry to solve the trimmed-down versions.
However, as the problem formulation was motivated by and done with\company{}, these constraints reflect an actual need from practice.
Further, we think that in the course of digitization in industry, companies will be faced with increasingly complex problems and solving them in an automated way without limitations might be a competitive advantage.

\section{Conclusion}
\label{sec:conclusion}
This work tackles the rich Vehicle Routing Problem~(rVRP) and its transfer to a real-world application.
We assess a multi-objective capacitated VRP with pickup and delivery~(PD) stops and time windows~(TW) and propose a two-staged strategy where the first step assigns the orders to the vehicles, and the second step optimizes the tours of each vehicle.
This diverse set of constraints delimits our work from other state-of-the-art approaches since these hardly cover a small set of these constraints.
We apply a six-dimensional cost function and propose a timeline algorithm to match the given TWs and fixed pause times.
To solve the problem instances on both stages, we apply a Genetic Algorithm~(GA) and Ant Colony Optimization~(ACO). 
We evaluate our approach on a real-world data set composed of eight different problem instances with increasing complexity in comparison to a Brute Force approach, two Blackbox algorithms provided by\company{}, and a Local Search algorithm.
Our evaluation has shown that our approach is able to tackle the defined rVRP and, hence, exceeds the functionality of the existing Blackbox algorithms. 
Further, it reduces the time to result compared to the existing algorithms by 90\% to two seconds for the TSP-stage and five seconds for the VRP-stage.
Therefore,\company{} already integrated our approach into their software and uses it actively.

In the future, we plan to investigate other common optimization algorithms such as particle swarm or Branch-and-Bound algorithms within our two-stage approach.
Further, we plan to examine whether a mixture of ACO, GA, and Local Search on the different stages might be beneficial. 
Also a multi-objective representation of the problem could be possible by transferring the constraints such as matching the time windows, reducing the required time and driving distance to be contrary objectives.
Then, we could apply common multi-objective optimization techniques such as NSGA-II and inspect their performance.
Following the observation from~\cite{fredericks2019planning}, that the selection of the algorithm for planning is situation-aware for adaptive systems, we also want to examine whether a situation-aware algorithm selection---for example based on the number of orders, vehicles, and drivers to assign---is meaningful for rVRP.
Finally, we think that integrating a measure of uncertainty or even forecasting mechanisms regarding orders or traffic could be beneficial, since orders cancelled at short notice or spontaneous full road closures due to accidents can upset the entire plan.

\bibliographystyle{IEEEtran}
\bibliography{Bibliography.bib} 

\end{document}